\documentclass{article}

\usepackage{microtype}
\usepackage{amsmath}
\usepackage{amssymb}
\usepackage{graphicx}
\usepackage{overpic}
\usepackage{bm}
\usepackage{subfigure}
\usepackage{booktabs} 
\usepackage{comment}
\usepackage{hyperref}

\usepackage[accepted]{icml2025}

\usepackage{amsmath}
\usepackage{amssymb}
\usepackage{mathtools}
\usepackage{amsthm}
\usepackage[normalem]{ulem}

\usepackage[capitalize,noabbrev]{cleveref}

\theoremstyle{plain}

\theoremstyle{definition}

\theoremstyle{remark}

\usepackage[textsize=tiny]{todonotes}

\newcommand{\figpanel}[1]{(\textbf{\lowercase{#1}})}

\newcommand\1{\leavevmode\hbox{\rm \small1\kern-0.35em\normalsize1}}
\newcommand\ind[1]{\1_{\{#1\}}}

\newcommand\indnew[1]{\1_{x}^{#1}}     

\newcommand\egaldef{\stackrel{\mbox{\upshape\tiny def}}{=}}
\newcommand\E{{\mathbb E}}
\newcommand\LL{{\mathcal L}}
\newcommand\I{{\mathbb I}}
\newcommand\C{{\mathcal C}}
\newcommand\G{{\bf G}}
\newcommand\n{^{\scriptscriptstyle (M)}}
\newcommand\Tr{\text{Tr}}
\newcommand\x{{\bm x}}
\newcommand\z{{\bm z}}
\newcommand\bC{{\bm C}}
\newcommand\J{{\bm J}}

\usepackage{xcolor}
\definecolor{myblue}{RGB}{12, 12, 158}
\definecolor{myred}{RGB}{158, 19, 22}
\definecolor{myorange}{RGB}{245, 150, 12}
\definecolor{mygreen}{RGB}{26, 148, 49}
\definecolor{Prune}{RGB}{99,0,60}
\definecolor{Purple}{RGB}{75, 0, 130}
\definecolor{Pink}{RGB}{255, 105, 180}
\definecolor{deepskyblue}{RGB}{0, 191,255}
\definecolor{limegreen}{RGB}{50, 205, 50}
\definecolor{crimson}{rgb}{0.86, 0.08, 0.24}

\definecolor{blue(ncs)}{rgb}{0.0, 0.53, 0.74}

\newcommand{\bx}[0]{{\bm{x}}}

\newcommand{\bJ}[0]{\bm{J}}

\newcommand{\D}[0]{{\mathcal{D}}}
\newcommand{\der}[0]{{\mathrm{d}}}

\newcommand{\like}[0]{{\mathcal{L}}}
\newcommand{\pemp}[0]{{{\mathcal{D}}}}

\newcommand{\caja}[1]{\left[ {#1} \right] }
\newcommand{\paren}[1]{\left( {#1} \right) }
\newcommand{\lazo}[1]{\left\{ {#1} \right\} }
\newcommand{\av}[1]{\left\langle {#1} \right\rangle }

\newcommand{\mean}[2]{\mathbb{E}_{{#2}}\left[ {#1} \right] }

\newcommand{\eigvecsymbol}{u}
\newcommand{\Cpop}[0]{{ {\bm C}^* }}
\newcommand{\Cpopval}[0]{{c^*_{\alpha}}}
\newcommand{\Cpopvec}[0]{{\boldsymbol{\eigvecsymbol}^*_{\alpha}}}

\newcommand{\CM}[0]{{{\widehat{\boldsymbol{C}}^{M}}}}
\newcommand{\CMval}[0]{{\hat{c}^{M}_{\alpha}}}
\newcommand{\CMvec}[0]{{\boldsymbol{\eigvecsymbol}^{M}_{\alpha}}}

\newcommand{\Jtrue}[0]{{\boldsymbol{J}^*  }}
\newcommand{\Jest}[0]{{\hat{\bm  J}}}
\newcommand{\JML}{{\hat{\boldsymbol{J}}^{\mathrm{ML},M}}}

\newcommand{\Cvalpop}[0]{{\widehat{\bm C}^{M}_{\text{val-pop}}}}

\newcommand{\CCorrectedRegu}[0]{\widehat{\bm C}^{M}_{\text{val}-L_2(\lambda)}}

\definecolor{mygreen}{RGB}{26, 148, 49}

\icmltitlerunning{A theoretical framework for overfitting in energy-based modeling}

\begin{document}

\twocolumn[
\icmltitle{
A Theoretical Framework For Overfitting In Energy-based Modeling}



\icmlsetsymbol{equal}{*}

\begin{icmlauthorlist}
\icmlauthor{Giovanni Catania}{ucm}
\icmlauthor{Aurélien Decelle}{ucm,politecnica}
\icmlauthor{Cyril Furtlehner}{paris}
\icmlauthor{Beatriz Seoane}{ucm}
\end{icmlauthorlist}
\icmlaffiliation{ucm}{Departamento de Física Teórica, Universidad Complutense de Madrid, Spain.}
\icmlaffiliation{paris}{Inria-Saclay, Université Paris-Saclay, LISN, Gif-sur-Yvette, France.}
\icmlaffiliation{politecnica}{Escuela Técnica Superior de Ingenieros Industriales, Universidad Politécnica de Madrid, Spain}
\icmlcorrespondingauthor{Giovanni Catania}{gcatania@ucm.es}
\icmlkeywords{energy-based models, overfitting, inverse problems, Bolzmann Machine, early-stopping, training dynamics, generative models}
\vskip 0.3in
]


\printAffiliationsAndNotice{}  
\begin{abstract} We investigate the impact of limited data on training pairwise energy-based models for inverse problems aimed at identifying interaction networks. Utilizing the Gaussian model as testbed, we dissect training trajectories across the eigenbasis of the coupling matrix, exploiting the independent evolution of eigenmodes and revealing that the learning timescales are tied to the spectral decomposition of the empirical covariance matrix. We see that optimal points for early stopping arise from the interplay between these timescales and the initial conditions of training. Moreover, we show that finite data corrections can be accurately modeled through asymptotic random matrix theory calculations and provide the counterpart of generalized cross-validation in the energy based model context. Our analytical framework extends to binary-variable maximum-entropy pairwise models with minimal variations.
These findings offer strategies to control overfitting in discrete-variable models through empirical shrinkage corrections, improving the management of overfitting in energy-based generative models.
Finally, we propose a generalization to arbitrary energy-based models by deriving the neural tangent kernel dynamics of the score function under the score-matching algorithm.
\end{abstract}

\section{Introduction}
Controlling overfitting is basic in machine learning, particularly as modern, over-parameterized architectures enhance learning capabilities. To prevent learning noise or irrelevant patterns, numerous empirical solutions have been proposed that modulate the model's implicit bias through its architecture and optimization, employing various implicit regularization mechanisms~\cite{belkin2021fit}. Finding the optimal balance between maximizing data utility, preserving generalization, and ensuring the privacy of training data represents a critical trade-off that can be challenging to pinpoint.
In supervised learning tasks like classification, overfitting is readily identified using standard practices. Metrics like test-set accuracy, particularly when augmented by cross-validation in data-scarce scenarios, clearly signal overfitting, enabling strategies like early stopping, regularization, and hyperparameter tuning to mitigate it. 
Furthermore, training and generalization performance in regression and classification tasks are now well understood in certain simplified regimes, such as high-dimensional ridge ~\cite{atanasov2024scaling,advani2020high,saxe_ganguli_dynamics_linearnetworks,tomasini_sclocchi_wyart_ridge_regression} or logistic~\cite{mai2019large,10.5555/3692070.3693394} regression
or numerous  more complex setting of non-linear regression in various scaling regime (see for instance ~\cite{mei2018mean,arnaboldi2023high,saad1995dynamics} among many other recent works). This gives the possibility to assess some simple indicator like the generalized cross-validation (GCV)~\cite{golub1979generalized}, an exact relation between train$\slash$test errors valid for the ridge regression that can be derived using a leave-one-out argument (see e.g.~\cite{cyril_regression}). This methodology is also relevant in deep learning contexts~\cite{wei2022more}, particularly in over-parameterized regimes where it aligns with  observed   stochastic gradient descent behaviors~\cite{patil2024failures}.

Recent advancements  in training and architecture have greatly enhanced the generative capabilities of neural network models across various fields~\cite{bengesi2024advancements}, enabling the creation of photorealistic images, credible speech synthesis, and biologically functional synthetic proteins~\cite{wu2021protein}. Despite this progress, selecting optimal models from a pool remains challenging due to noisy training data often leading to undetected overfitting. Yet, detecting overfitting in unsupervised learning settings, particularly for generative modeling, is elusive but crucial, especially with sensitive datasets like human genomic data~\cite{yelmen2021creating,yelmen2023deep} and copyrighted content. Unlike supervised learning, unsupervised learning lacks clear overfitting indicators, complicating model development and validation. While some theoretical insights on optimal regularization tuning for simple energy-based models exist~\cite{Fanthomme_2022}, practical indicators such as early stopping points during training dynamics remain undefined. Moreover, estimating log-likelihood for model selection poses significant computational challenges~\cite{bereux2022learning,bereux2024fast}. Consequently, there is an urgent need for methods to detect and mitigate overfitting in these contexts.

This paper focuses on energy-based models (EBMs)~\cite{ackley1985learning}, which encode the empirical distributions of various data types ---such as neural recordings~\cite{roudi2009statistical}, images~\cite{du2019implicit}, and genomic~\cite{yelmen2021creating} or proteomic~\cite{morcos2011direct} sequences---into a probability framework rooted in Boltzmann's law. By adopting a Bayesian approach, EBMs aim to maximize the likelihood function, enabling the generation of new data that closely resembles the training set and facilitates the extraction of detailed microscopic insights.
EBMs range from simple Boltzmann Machines (BMs) and Restricted Boltzmann Machines (RBMs) to more complex architectures like convolutional neural networks, making them versatile in statistical physics for solving inverse problems like deducing Hamiltonian parameters from observed data. The interpretability of simple EBMs enables to uncover underlying rules within datasets: this capability has proven highly effective in fields ranging from neuroscience~\cite{roudi2009statistical} to bio-molecular structure prediction~\cite{cocco2018inverse} predominantly through pairwise maximum-entropy models. Recent advancements extend these applications, using complex EBMs to infer high-order interactions~\cite{RBMeffectivemodel_ising,RBMeffectivemodel_potts,feinauer2022interpretable,feinauer2022reconstruction} or constitutive patterns~\cite{tubiana2019learning,decelle2023unsupervised}, significantly deepening our comprehension of data structures.

This work develops a theoretical framework for understanding and mitigating overfitting in EBMs. We begin with a simple Gaussian model as a fundamental non-trivial example, using it to quantitatively analyze overfitting through synthetic experiments with predefined ground truths. We examine eigenvalue dynamics using artificial covariance matrices that simulate real datasets, exploring how overfitting arises from different learning timescales associated with various eigenmodes of the empirical covariance matrix. We address inaccuracies in learned eigenvalues with corrections based on random matrix theory (RMT), showing that the quality of model generation in EBMs is less affected by the lower modes of the covariance matrix, while the accuracy of inferred couplings is significantly impacted. We demonstrate that regularization techniques like shrinkage corrections are crucial to counteract overfitting, providing a robust framework to refine EBM training by considering finite-sample-size effects. This approach also informs our analysis of more complex models like the BM, underscoring the importance of  regularization strategies to enhance model reliability and predictive accuracy. \section{Gaussian Model} The Gaussian Energy-Based Model (GEBM) specifies a multivariate Gaussian distribution for real-valued variables \(\bm{x} \in \mathbb{R}^N\), characterized by 2-body interactions encoded within a symmetric, positive-definite coupling matrix, \(\bm{J} \in \mathbb{R}^{N \times N}\). The GEBM is the simplest model that effectively captures the first and second-order statistics of a set of data. For the purposes of this analysis, we assume $0$ means for the data components, thus simplifying the initial model by excluding the learning of external biases. Nevertheless,  the theoretical framework presented below can be readily extended to the above to accommodate non-zero means.  The probability distribution of a configuration $\bm x$ is then:\begin{equation}
    \textstyle p\left(\bm x \! \mid\! \bm J \right) \!=\! \left(2\pi \right)^{-N\slash 2} \sqrt{\det \bm J} e^{-\frac{1}{2} \bm x^\top \bm J \bm x}.\label{eq:GEBM_pdf}
\end{equation}
It is straightforward to check that the population covariance matrix of such distribution is $\bm C = \mean{\bm x \bm x^\top}{\bm J}=\bm J^{-1}$, with $\mean{\cdot}{\bm J}$ denoting the average with respect to \eqref{eq:GEBM_pdf}.

\textbf{Inference problem.}
Consider a dataset \(\mathcal{D} = \{\bm{x}^{\mu}\}_{\mu=1}^M\) with \(M\) entries generated with a GEBM model with coupling matrix $\Jtrue$.
Our objective is then to find the parameters $\Jest$ that best approximate the empirical distribution of the data ---formally $p_\mathcal{D}(\bm x)=M^{-1}\sum_{\mu=1}^M\delta \paren{\bm x-\bm x^{\mu}}$---, 
with the probabilistic model~\eqref{eq:GEBM_pdf}. Without prior information about the model parameters \(\bm J\), the maximum likelihood (ML) estimator \(\JML\) is calculated as \(\JML = \paren{\CM}^{-1} \), where $\CM$ is the empirical covariance matrix from \(M\) data points, provided it is invertible \cite{MacKay2003}. Denoting with $N$ the number of data components (data dimensions), this condition requires that $M\geq N$, assuming samples to be independent. Clearly, when $M\to\infty$, $\JML$ recovers the true set of parameters used to generated the data, $\Jtrue$.

\textbf{Training dynamics.}
The GEBM stands out as one of the few high-dimensional inference problems where an analytical expression for the ML estimator is available, independent of both $M$ and $N$. However, our focus here is on the training dynamics associated with an iterative maximization of the likelihood function through gradient ascent dynamics, as is typical in EBMs. This approach allows us to explore the adaptive process of parameters' estimation over time.

In the GEBM, the log-likelihood (LL) of the parameters \(\bm{J}\) depends only on $\CM$ and it reads $
   \textstyle \like (\bm J) = -\frac{1}{2}\sum_{i,j} J_{ij} \widehat{C}_{ij}^M + \frac{1}{2} \log \det \bm J\label{eq:likelihood_gaussianmodel}
$.
This quantifies how well \(\bm{J}\) matches the observed data.  In a standard gradient ascent algorithm, the update rule for the parameters reads: \begin{equation}
   \textstyle J_{ij}^{t+1} =  J_{ij}^{t} + \gamma \frac{\partial \like}{\partial J_{ij}},\label{eq:gradient_ascent}
\end{equation}
where $\gamma$ is the learning rate. Assuming non-symmetric perturbations on the parameters $J_{ij}$, the gradient in \eqref{eq:likelihood_gaussianmodel} reads: 
\begin{equation}
 \textstyle \!\! \frac{\partial \like}{\partial J_{ij}}\! =\textstyle\!  \textstyle\! - \widehat{C}^M_{ij} \!+\! \mean{x_i x_j}{\bm J} \!=\textstyle\! \!\textstyle - \widehat{C}^M_{ij}\! +\! \left(\bm J^{-1}\right)_{ij}.\label{eq:gradient_gaussian_model}
\end{equation} The second equality comes from the exact expression of 2-point correlations of the Gaussian model in terms of its coupling matrix $\bm J$: this is equivalent to assume that we perfectly sample the model with an infinite amount of configurations at any $t$. For a more generic EBM, another source of noise should be added due to the finite number of samples (or chains) used to estimate the empirical correlations used to compute the gradient. 

To analyze the learning dynamics, we use the spectral decomposition of \(\bm{J}\) and project the gradient onto its eigenbasis, denoted by \(\bm{V} = \{\bm{v}_\alpha\}_{\alpha=1}^N\). From \eqref{eq:gradient_gaussian_model}, the gradient projected on modes $\alpha$ and $\beta$ is expressed as:
\begin{equation}
    \textstyle \left(\frac{\partial \like}{\partial \bm J} \right)_{\alpha \beta}= -{\hat{c}}^M_{\alpha \beta} + \frac{\delta_{\alpha \beta}}{J_{\alpha}}, \label{eq:projected_LL_gradient}
\end{equation} 
 where \(\hat{c}_{\alpha \beta}^M\) is the projection of $\CM$. Generally, for any matrix \(\bm{\mathcal{M}}\), we write \(m_{\alpha \beta}\egaldef\bm{v}_\alpha^\top \bm{\mathcal{M}} \bm{v}_\beta\). 
 
 This approach enables us to formulate a set of evolution equations for the eigenvalues \(\lbrace J_\alpha \rbrace\) and the rotation of the eigenvectors. By assuming an infinitesimal learning rate, we can transform the discrete-time update equation~\eqref{eq:gradient_ascent} into a continuous set of differential equations (see Appendix~\ref{app:derivation_gradient} for further derivation details):
 \begin{align}
      \!\!\tau\frac{\der J_\alpha}{\der t} \!=\! \frac{1}{J_\alpha}\!-\!\hat{c}_{\alpha \alpha}^M; \quad &\:\:\:\:\:\:
     \tau\bm{v}^\alpha \frac{\der\bm{v}^\beta}{\der t}\! =\! \frac{\hat{c}_{\alpha \beta}^M}{J_\alpha \!-\! J_\beta}\!\text{ for }\alpha\!\neq\!\beta,\!\! \label{eq:grad_Ja}
\end{align}
where $\tau$ is a timescale set by the learning rate, $\tau=1\slash \gamma$. From Eq.~\eqref{eq:grad_Ja} we see that eigenvectors of $\bm J$ stop rotating when they align with the eigenvectors of $\CM$ to cancel out the numerator $\hat{c}_{\alpha \beta}^M$. Eq.~\eqref{eq:grad_Ja} can be integrated analytically, and $\hat{c}_{\alpha\alpha}^M$ can be replaced by the matrix eigenvalue $\hat{c}_{\alpha}^M$.  

The solution of \eqref{eq:grad_Ja} can be expressed in an explicit (although not closed) form using Lambert $W_0$ function, namely:
 \begin{equation}
     J_{\alpha} \paren{t} = \frac{1}{\hat{c}_{\alpha}^M} + \frac{1}{\hat{c}^M_{\alpha}} W_0 \caja{\mathcal{B}_\alpha e^{- \paren{\hat{c}^M_{\alpha}}^2 \frac{t}{\tau}}},\label{eq:JalphatGaussianEBM_lambert}
 \end{equation}
with the constant $\mathcal{B}_\alpha$ is fixed by the initial condition at $t=0$.  
Eq.~\eqref{eq:JalphatGaussianEBM_lambert} delineates the evolution of each eigenvalue, which progresses independently once the eigenvectors of \(\bm{J}\) align with those of $\CM$. A crucial aspect of this equation is that the relaxation time it takes for an eigenvalue to reach its steady-state value $J_{\alpha}^{(\infty)} \!=\! \lim_{t \to \infty} J_{\alpha}(t) \!=\! 1 / \CMval $ is inversely proportional to the square of the corresponding eigenmode in the covariance matrix: indeed, for $t\!\to\!\infty$ Eq.~\eqref{eq:JalphatGaussianEBM_lambert} describes an exponential relaxation to the fixed point with a timescale \(\propto (\CMval)^{-2}\).\\ This relationship shows that the evolution of each eigenvalue is closely linked to the significance of the corresponding eigenvector in representing the data, so that stronger modes in the covariance matrix are learned more quickly than weaker ones. 
The idea that information is learnt progressively starting from strong PCA's directions is closely related to the concept of spectral bias \cite{spectralBiasBengio} -  although here the decomposition is spectral rather than in a Fourier basis - and it has been characterized theoretically in the case of linear regression~\cite{advani2020high}. The interaction between these varying timescales can result in an initial phase where the strongest components of the dataset's PCA are effectively captured, followed by a phase where training begins to adjust noise-dominated directions, potentially leading to overfitting.

We've shown that GEBMs' learning dynamics are governed by the spectral decomposition of $\CM$, with finite-sample effects arising from changes in the spectrum due to finite \(M\). This falls within the realm of random matrix theory (RMT)~\cite{potters2020first}, as we  detail shortly.

\textbf{Asymptotic RMT analysis.}
In our simplified setting of GEBMs, the parameters of the model $\J(t)$ along the learning 
trajectory are an explicit function of the empirical covariance matrix $\CM$ upon choosing the same constant 
${\mathcal B}_\alpha = {\mathcal B} > -1/e$ in~\eqref{eq:JalphatGaussianEBM_lambert} for the initialization ($\mathcal{B} = -1/e$ corresponds to the initialization $\J(0) = 0$) and assuming 
that $\bm J (t)$ is aligned with $\CM$ at $t=0$. This choice simplifies considerably the analysis, and as explained in Appendix~\ref{sec:RMT}. Using RMT, all relevant quantities can be derived in closed forms based solely on the {\em population spectrum} \(\nu\) and the {\em aspect ratio} \(\rho = M/N\), under the asymptotic proportional scaling where \(M, N \to \infty\) with \(\rho\) held constant.

We are interested in the train and test energies:
\begin{align}
    \textstyle E_{\rm train} \!=\! N^{-1}\Tr[\J \CM]\:\text{ and }\: E_{\rm test} \!= \!N^{-1}\Tr[\J \Cpop],\nonumber
\end{align}
 the coupling error $\textstyle {\mathcal E}_{\rm J} \!\egaldef\! N^{-1}\Vert \J-\Jtrue\Vert_F^2 $, with $\Vert\cdot\Vert_F^2$, the Frobenius norm, and the LL (train and test)
\begin{align}
\textstyle LL_{\rm train,test} &\egaldef\textstyle \frac{1}{2N}\log\det[\J]-\frac{1}{2}E_{\rm train,test}\label{def:LL}
\end{align}
where $\Cpop \egaldef \lim_{M\to\infty} \CM$ 
is the population matrix. In addition, we will also explore the behavior of the maximizer of the log-likelihood with regularization, i.e.
$\textstyle\LL[\J] = LL_{\rm train}[\J] -\lambda A(\J)$
focusing on \(A(\J)=\Tr[\J^2]\) for \(L_2\) ridge regularization, and \(A(\J)=\Tr[\J]\) for \(\widetilde{L}_1\) lasso regularization on the spectrum. $\widetilde{L}_1$ is applicable since \(\J\) is symmetric and remains positive definite throughout the trajectory, ensured by the logarithmic barrier.

We simply quote here the result of the asymptotic limits  (for $\rho\!>\!1$, more details in Appendix~\ref{sec:RMT}), based on RMT~\cite{MaPa,ledoit2011eigenvectors}. First, the spectral density $\bar\nu$ of $\CM$ reads in this limit:
\begin{equation}\label{eq:bnu}
\bar\nu(x) = \frac{\rho \Lambda_i(x)}{\pi x} = \textstyle\frac{\rho}{\pi x}\textstyle \frac{\Gamma_i(x)}{\bigl[1-\Gamma_r(x)\bigr]^2+\Gamma_i(x)^2},
\end{equation}
where $\Lambda(z) = \Lambda_r(x)+ i\Lambda_i(x)$ and $\Gamma(z) = \Gamma_r(x)\pm i\Gamma_i(x)$
for $z=x+i0^+$,  obey the self-consistent equations
\[
 \textstyle \Lambda(z) = \frac{1}{1-\Gamma(z)\tau},\qquad
 \textstyle \Gamma(z) = \frac{1}{\rho}\int \frac{\nu(dx) x}{z-\Lambda(z) x},
\]
in terms of the population spectral density $\nu(dx)$. In turn we obtain 
\begin{align*}
\textstyle E_{\rm train} &= \textstyle\frac{\rho}{\pi}\int_0^\infty dy j(y) \bigl[\Lambda_r(y)\Gamma_i(y)+\Lambda_i(y)\Gamma_r(y)] ,\\[0.2cm]
E_{\rm test} &=\textstyle \frac{\rho}{\pi}\int_0^\infty dy j(y) \Gamma_i(y), \qquad \quad\quad (\rho\ge 1),
\end{align*}
while the coupling error takes the form
\[
{\mathcal E}_{\rm J} = \textstyle\int_0^\infty\frac{\nu(dx)}{x^2}+\int_0^\infty \bar\nu(dx)j(x)\Bigl[j(x) -\frac{2}{\rho}\frac{(1-\rho)+2\rho\Lambda_r(x)}{x}\Bigr]
\]
where $j(x)$ is one of the analytical functions $j_t,j_{L_1}$ and $j_{L_2}$ corresponding respectively to the time dependent, $\widetilde{L}_1$ and $L_2$ regularized forms of $\J$. Remarkably, for the  $\widetilde{L}_1$ regularized coupling matrix we get a deterministic relation between the train and test energies (Appendix~\ref{sec:RMT}) \begin{equation}
    E_{\rm test} = \paren{1-\rho^{-1} E_{\rm train}}^{-1} E_{\rm train},
\end{equation}
which is the counterpart of GCV for GEBMs which might be usable in practice for arbitrary EBM (in the same way as GCV can be used for deep regression models), as it allows one to get an estimation of the test LL. This concept remains a topic for future research. Instead, we have focused on strategies for data cleaning and regularization, specifically employing shrinkage techniques~\cite{bouchaud_cleaning}.
Using a model of the data defined in the next section, allows us to specify $\nu(x)$ in order to assess
these strategies by comparing with the expected optimal performances given by RMT. 
\begin{figure}[t]
\centering
\begin{overpic}[width=\columnwidth,trim=0 30 0 0]{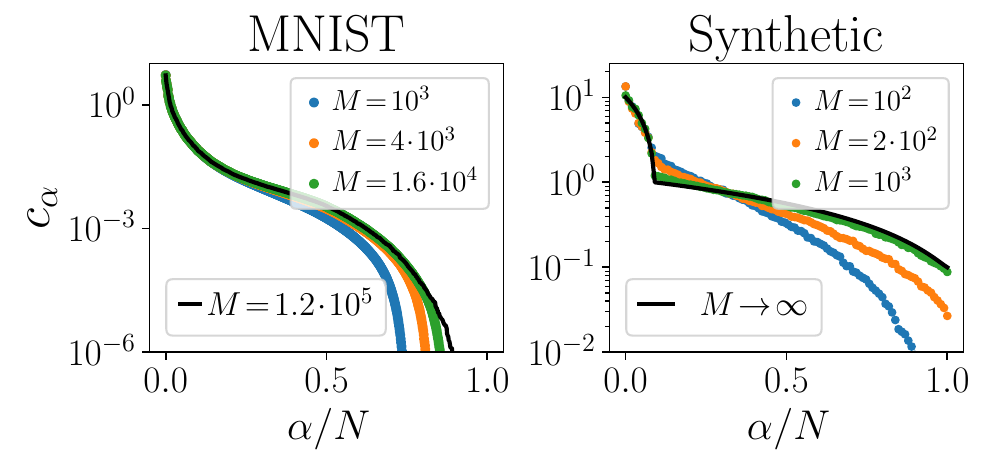}
\put(9,37.5){{\figpanel{a}}}
\put(55,37.5){{\figpanel{b}}}
\end{overpic}
\caption{\figpanel{a}: Eigenvalue spectra of the empirical covariance matrices for MNIST dataset~\cite{deng2012mnist}. Black lines show spectra using the full dataset size (\(M^*\)), while scatter colored points represent subsets (\(M\! <\! M^*\)). \figpanel{b}: Black line shows a synthetic population eigenvalue spectrum based on \eqref{def:Cpopmodes} for \(N\!=\!100\), \(r\!=\!0.9\), \(\beta\!=\!0.9\), \(\gamma\!=\!1.1\), \(x_1\!=\!10^{-1}\), \(x_2\!=\!10\); colored points show the eigenvalues from \(\CM\) calculated by sampling different \(M\) configurations from a GEBM model with \(\Jtrue \!=\! \Cpop^{-1}\) (Eq. \eqref{eq:GEBM_pdf}).\label{fig:eigenvalues_realandsynthetic}}
\end{figure}

\section{Modeling realistic data covariances}
To effectively study the impact of finite number of data on the learning process of a GEBM in a controlled setting, we need to define a synthetic model that facilitates the analysis of different learning timescales. The first step is to artificially create a \textit{population} covariance matrix \(\Cpop\), from which a ground truth coupling matrix \(\Jtrue\) is constructed, through \(\Jtrue = \Cpop^{-1}\). Using this setup, we generate a multivariate Gaussian distribution and extract \(M\) data points from it. These data points are then used to train a new GEBM using the \textit{empirical} covariance matrix, \(\CM\), derived from the \(M\) of these samples, with the goal of inferring the original model parameters.

As previously discussed, the training dynamics of each mode of \(\bm J\) are directly linked to the eigenvalues of \(\CM\). To enhance this analysis, we have developed a synthetic model for the spectrum of \(\Cpop\), which influences the spectrum of \(\CM\) in scenarios with finite datasets. This model closely mimics the eigenvalue spectra of real datasets, as illustrated in Figure~\ref{fig:eigenvalues_realandsynthetic}-\figpanel{a}, which shows the eigenvalue spectrum (in descending order) of covariance matrices from MNIST for several sizes \(M\) (more examples are given in Appendix~\ref{app:eigenvectors}).
Our analysis reveals that the spectrum of \(\CM\) remains relatively stable w.r.t. $M$ for a significant number of modes, indicating \(\hat{c}_\alpha^M \! \approx \! c_\alpha^\infty\ \!=\! \Cpopval \). However, smaller eigenvalues fluctuate markedly with \(M\); they tend to be underestimated as \(M\) decreases, suggesting \(\CMval < \Cpopval\) for small \(\Cpopval\), and are slightly overestimated for the larger eigenvalues. This behavior is rigorously characterized using RMT tools in simplified data models~\cite{baik2006eigenvalues,ledoit2011eigenvectors}. Additional insights into the conservation of eigenvectors across modes are detailed in Appendix \ref{app:eigenvectors}.

Inspired by these findings, we will characterize our synthetic population matrix \(\Cpop\) by an eigenvalue spectrum \(\{\Cpopval\}_{\alpha=1}^N\) generated according to a mixture of power laws. The cumulative distribution is defined as follows:

\begin{equation}
    \textstyle \!\!P\!\caja{\lambda \!<\! x} \!=\! r \!\caja{\frac{x\!-\!x_1}{1\!-\!x_1}}^{\!\beta}\!\!\!\indnew{(x_1,1)}\!\!+\!\! \left[r\!+\!\textstyle\paren{1\!-\!r}\!\textstyle\!\paren{\frac{x-1}{x_2-1}}^{\!\!\gamma}\right]\!\indnew{(1,x_2)}\!\!\label{def:Cpopmodes}
\end{equation} 
where $\indnew{(a,b)}$ denotes the indicator function in the interval $(a,b)$. This setup distinguishes between ``strong" modes with \(\Cpopval > 1\) and ``weak" modes with \(\Cpopval < 1\), with their prevalence controlled by parameter \(r\). Parameters \(\beta\) and \(\gamma\) represent the power-law exponents for these two categories, with \(x_1\) and \(x_2\) the respective lower and upper cutoffs. Model \eqref{def:Cpopmodes} is chosen to $i)$ mimic the eigenvalue distribution of a realistic dataset's covariance matrix, though our numerical results are robust to specific spectral details, and $ii)$ to extract asymptotic quantities in the continuous-density limit \(N,M \to \infty\) (with $\rho=M/N$ finite) using RMT.

Figure~\ref{fig:eigenvalues_realandsynthetic}-\figpanel{b} shows an example spectrum of population eigenvalues \(\Cpopval\) from Eq.~\eqref{def:Cpopmodes} with \(N=100\) (black line), alongside empirical estimates of finite-data eigenvalues \(\CMval\) (scatter points) for various \(M\) values, demonstrating that strong modes remain stable despite finite-\(M\) noise, while weak modes are consistently underestimated. The full matrix \(\Cpop\) is finally assembled by projecting the diagonal matrix of these eigenvalues onto a random orthogonal matrix \(\bm U ^* = \{ \Cpopvec \}_{\alpha=1}^{N}\), resulting in \(\Cpop=\sum_{\alpha} \Cpopval \Cpopvec {\Cpopvec}^{\top}\).

\section{Training Dynamics on Synthetic Data\label{sec:training}}

\begin{figure}[t]
    \centering
\begin{overpic}[width=\columnwidth, trim = 0 30 0 0]{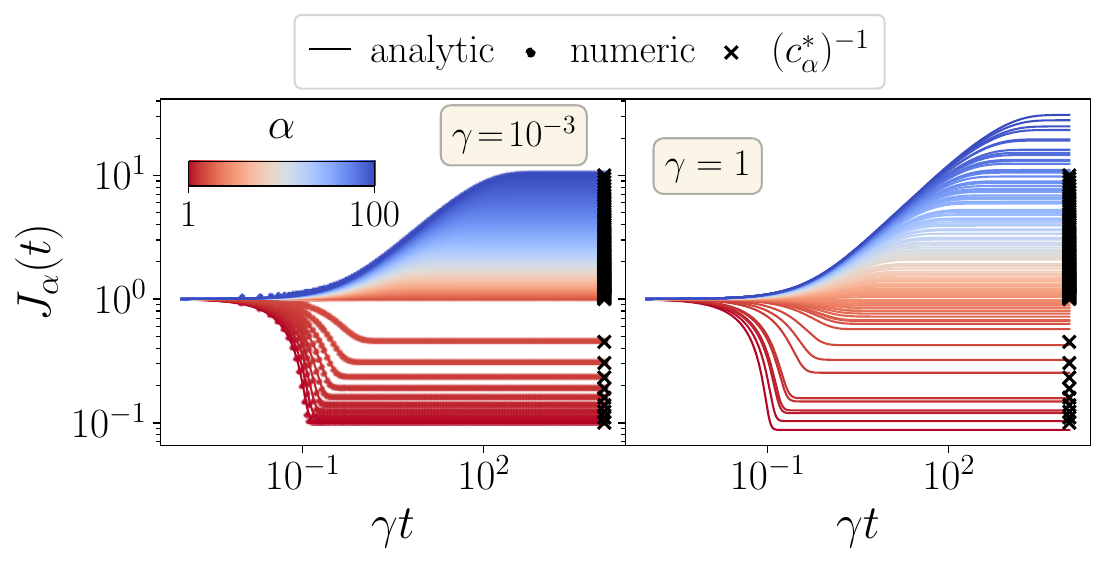}
\put(16,7){{\figpanel{a}}}
\put(59,7){{\figpanel{b}}}
\end{overpic}
\caption{
Training dynamics of the GEBM from a population matrix \(\Cpop\) (in \figpanel{a}, with system size and parameters matching those in Fig.~\ref{fig:eigenvalues_realandsynthetic}-\figpanel{b}), and from an empirical covariance matrix $\CM$ (generated from $\Cpop$ through \eqref{eq:likelihood_gaussianmodel}, with $\rho\!=\!2.11$ (in \figpanel{b}). \figpanel{a}-\figpanel{b} display the analytic evolution of eigenvalues \(J_{\alpha}\) toward the steady-state (lines), and a comparison with numerical training (points, in \figpanel{a}). In all cases the initial condition is an identity matrix. \label{fig:gauss_training_population}}
\end{figure}

The introduced synthetic model enables analysis of training dynamics in two scenarios: \(i)\) an ideal setting with an infinite amount of samples using \(\Cpop\) as the data covariance matrix, and \(ii)\) a more realistic situation with a finite dataset, represented by the empirical covariance matrix \(\CM\).

\textbf{Training Dynamics with Infinite Data.}
For clarity, we begin by training our GEBM using the population covariance matrix \(\Cpop\). Fig.~\ref{fig:gauss_training_population}-\figpanel{a} illustrates the evolution of the coupling matrix eigenvalues \(J_{\alpha}\), comparing analytical solutions from Eq.~\eqref{eq:JalphatGaussianEBM_lambert} and numerical iterative training using Eq.~\eqref{eq:gradient_ascent}, both starting from the same initial condition (\(J_{\alpha}(0)\!=\!1\)). The analytical and numerical results match perfectly, demonstrating the expected time-scale separation: eigenvalues \(J_{\alpha}\) corresponding to stronger covariance modes converge faster to their fixed point \(J_{\alpha}^{(\infty)}\! =\!  J^*_{\alpha} \!=\! 1/\Cpopval\), which are the smallest in the coupling matrix, while weaker modes converge slower. Starting from an initial condition \(\bm J (0)\) that does not commute with \(\Cpop\), the coupling matrix must initially align its eigenvectors with those of \(\Cpop\), a process detailed in Appendix~\ref{app:GEBM_eigenvectors} and guided by Eq.~\eqref{eq:grad_Ja}. Following this alignment, the eigenvalues evolve independently according to Eq.~\eqref{eq:Jalpha_lambert_Ising}, supporting our analytical approach. Notably, training directly from the population matrix achieves perfect reconstruction of the original model, thereby avoiding any discrepancies or generation errors as expected.

\begin{figure}[t!]
\begin{overpic}[width=\columnwidth, trim = 0 30 0 0]{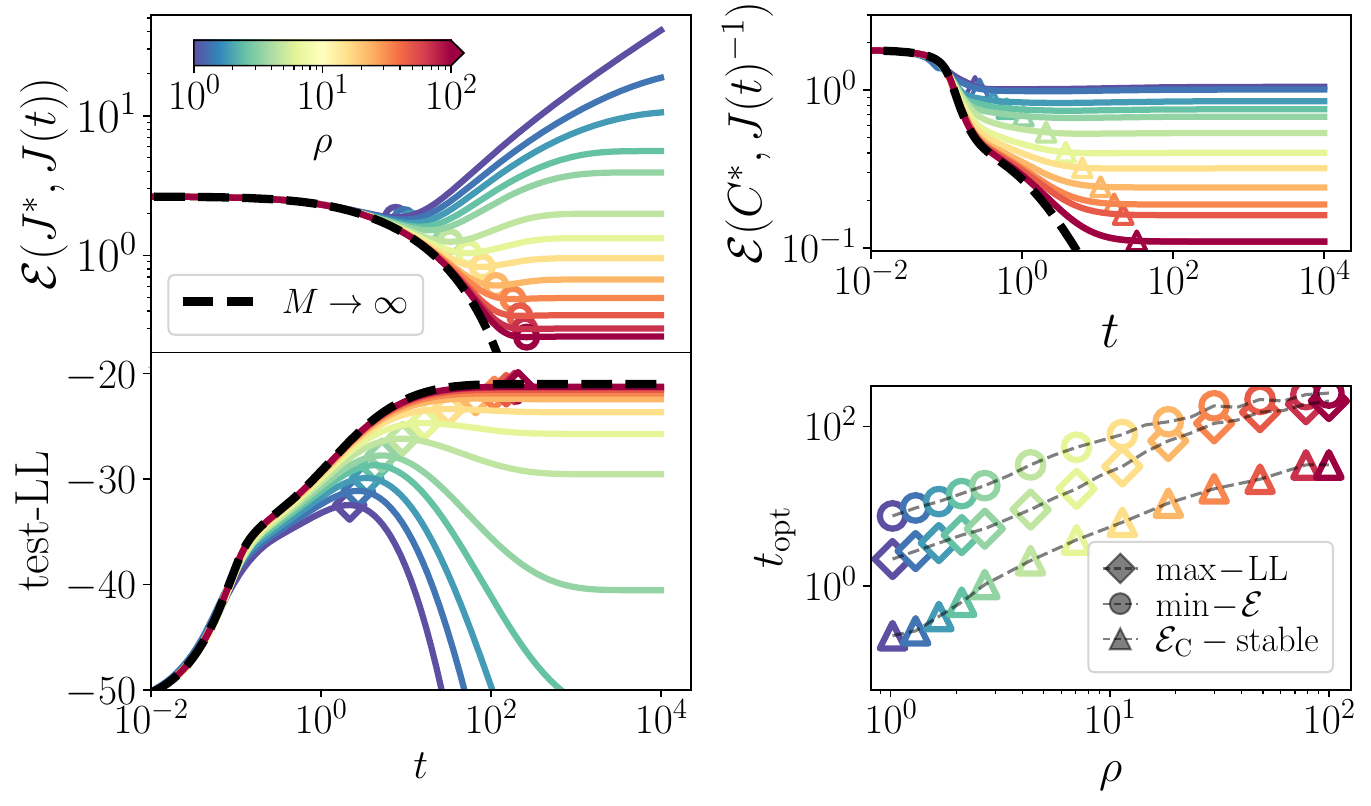}
\put(4,50.5){{\figpanel{a}}}
\put(13,22.5){{\figpanel{b}}}
\put(65,38.5){{\figpanel{c}}}
\put(65,22.){{\figpanel{d}}}
\end{overpic}
\caption{Results for GEBM training with finite data. \figpanel{a}-\figpanel{b}-\figpanel{c} display respectively the reconstruction error \(\mathcal{E}_\text{J}\), the test-LL and the generation error $\mathcal{E}_\text{C}$, all plotted vs time, for various sample sizes \(M\) (indicated by a color gradient from blue to red for increasing $\rho\!= \!M \!\slash\! N$. Dashed black lines refer to a training from \(\Cpop\) (i.e \(M \!\to \!\infty\)). 
\figpanel{d}: comparison between time of minimum reconstruction (circles), maximum test LL (diamonds) and time at which the generation error converges to its steady-state value. These quantities are also shown in the related panels for better clarity.
\label{fig:gaussian_model_differentM}}
\end{figure}

\textbf{Impact of Finite Datasets: Interplay of Initialization and Time Scales Favoring Early Stopping Strategies.}
We explore the training dynamics using finite-data estimates of the population covariance matrix, \(\CM\). With any finite \(M\), the  GEBM trained with \(\CM\) will show discrepancies from the true model \(\Jtrue\). We track these discrepancies by computing the reconstruction error $\mathcal{E}_{\text{J}}$ between \(\Jtrue\) and the trained \(\bm J(t)\) defined in the previous section. 
Fig~\ref{fig:gaussian_model_differentM}-\figpanel{a} illustrates the error's evolution over training time. Beginning from an identity matrix, at low \(\rho=M\!\slash\! N\) values, the error displays marked non-monotonic behavior, peaking at a specific \(t_{\mathrm{min}}(\rho)\) before stabilizing at the training's fixed point. At higher \(\rho\), the error decreases monotonically until stabilization, following a trend consistent with the \(M \!\to\! \infty\) scenario (i.e. using \(\Cpop\), in black dashed line). This behavior, also noted in complex EBMs \cite{RBMeffectivemodel_ising,nonconvergentsamplingICML}, underscores the GEBM's utility as a simple model yet capturing complex phenomena in EBMs.

This analysis shows that with limited data, there is an optimal training duration beyond which model inference accuracy declines, highlighting a sweet point for early stopping. However, detecting this point without ground truth is challenging: it does not coincide with the peak of test LL (as in \figpanel{b}), a phenomenon also noted in RBMs~\cite{RBMeffectivemodel_ising}. Moreover, the generation's quality, given error between \(\bm C(t) = \bm J(t)^{-1}\) and the population matrix $\Cpop$ in \figpanel{c} (computed as \(\mathcal{E}_{\text{C}}\! \egaldef\! \|\Cpop \!-\! \bm C(t)\|_\mathrm{F}\)), stabilizes well before \(t_{\mathrm{min}}(\rho)\) and remains flat afterwards. This suggests that the generation quality of the GEBM isn't solely dependent on the model itself, as evidenced by consistent generation errors at both the minimum-error point \(t_{\mathrm{min}}\) and the training's fixed point, indicating this metric fails to capture the deterioration of model parameters over time. These optimal times are shown against \(\rho\) in Fig~\ref{fig:gaussian_model_differentM}-\figpanel{d}. Additional evaluation metrics are discussed in Appendix~\ref{app:different_spectra_and_wasserstrain}.

Fig.~\ref{fig:gauss_training_population}-\figpanel{b} illustrates the evolution of eigenvalue \(J_{\alpha}\) over time for a sample size with \(\rho\!=\!2.11\). Initially, the stronger modes (deep red curves), which are less influenced by low-\(M\) induced noise, quickly stabilize, aligning their eigenvalues \(\CMval\) close to the population values \(\Cpopval\), marked by black crosses. This early alignment to very small and relatively accurate values makes the error curve $(\mathcal{E}_\text{J}$) for finite \(M\) closely resemble that of the \(M \!\to\! \infty\) scenario. However, after a period around \(t_{\mathrm{min}}(\rho)\), the model starts encoding the weaker modes (blue lines), which are systematically understated relative to the population, causing \(J_{\alpha}^{(\infty)}\! =\! 1 / \CMval\) to significantly exceed the ground-truth \(  J^*_{\alpha} \!=\! 1 / \Cpopval\). If training begins from small \(J_{\alpha}\) values, there's a critical point where the eigenvalues temporarily align more closely with their ground-truth than at the fixed point, effectively creating an optimal time $t^*_{\alpha}$ where \(J_{\alpha}(t^*_{\alpha}) \approx J^*_{\alpha}\). This alignment markedly decreases discrepancies between the trained model's eigenvalues and those of the true model, highlighting the significance of initial conditions in training dynamics. Yet, the specific initial values of \(J_{\alpha}(0)\) are less critical, as long as they are substantially smaller than \(1/\CMval\) for the weaker modes of $\CM$ (see Appendix~\ref{sec:other_inits} for further details).
\begin{figure}[t]
    \centering
\begin{overpic}[width=1\columnwidth, trim = 0 30 0 0]{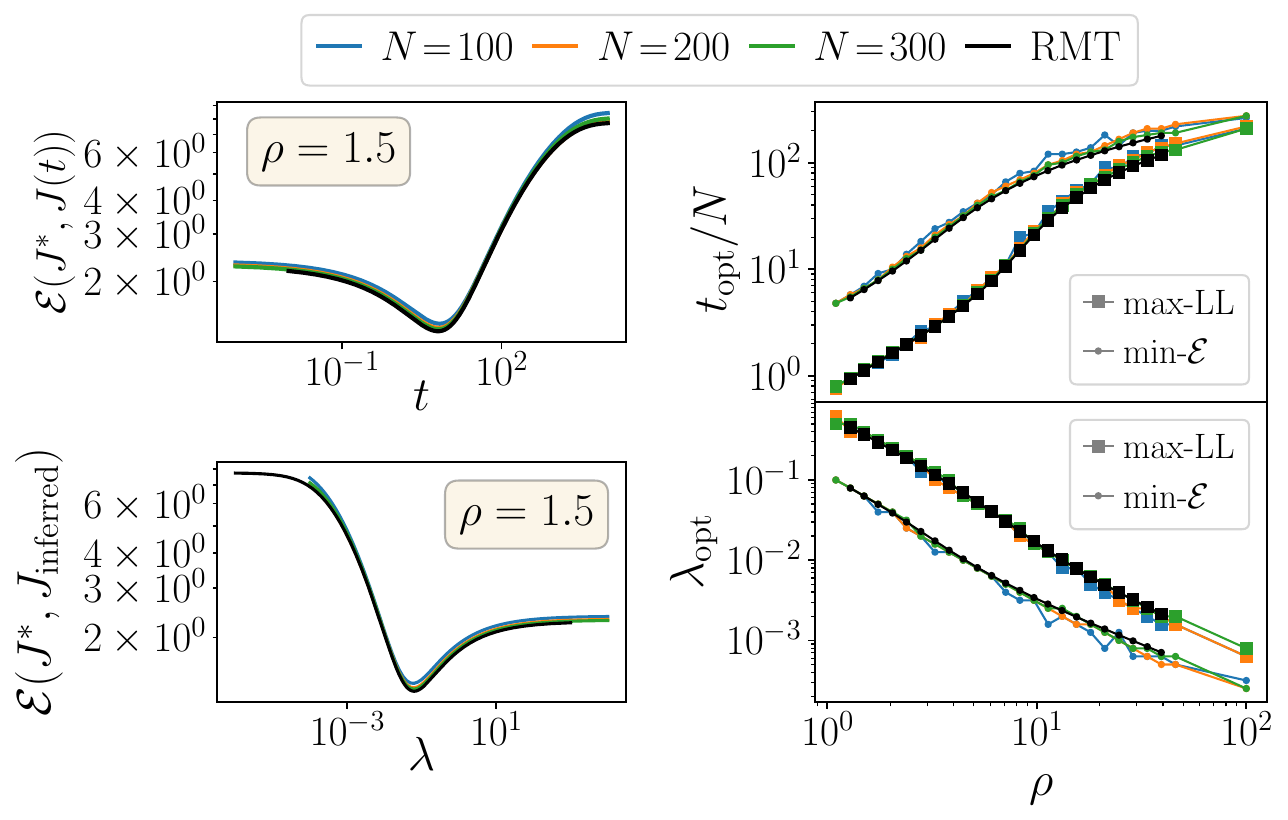}
\put(42,34.5){{\figpanel{a}}}
\put(65,46.5){{\figpanel{b}}}
\put(19,7){{\figpanel{c}}}
\put(65,7){{\figpanel{d}}}
\end{overpic}
\caption{\figpanel{a}-\figpanel{c}: for \(\rho\!=\!M/N\!=\!1.5\) we plot the reconstruction error during training (in \figpanel{a}, vs $t$) and the final reconstruction obtained using a $L_2$-norm regularization (in \figpanel{c}, vs $\lambda$).
\figpanel{b}: training time achieving optimal reconstruction error (points) and time of maximum test LL (squares), plotted vs $\rho$. \figpanel{d}: optimal value of regularization prior $\lambda$ vs $\rho$, again selecting the optimum w.r.t. reconstruction error and w.r.t. the test LL. All panels show comparisons between numerical results for various \(N\) (colored lines) against asymptotic results from RMT (black line).
\label{fig:rmt_comparison_dynamics}}
\end{figure}

Now, we can also explain the stable generation performance of the GEBM, shown in Fig.~\ref{fig:gaussian_model_differentM}-\figpanel{c}, using scale separation arguments. Generation error mainly depends on the strongest $\CMval$ values, which are learned early on, whereas the overall model quality is controlled by the weakest $\CMval$ (where \(J_{\alpha}^{(\infty)} \!=\! 1/\CMval\)), which minimally affects generation error due to their small value. While this phenomenon appears unique to the GEBM, a similar effect is observed in binary pairwise EBMs (cf. Sec.~\ref{sec:BM}).

\textbf{Asymptotic analysis.} Our findings so far have been established by numerically integrating the gradient ascent dynamics (i.e. using Eq.~\eqref{eq:gradient_ascent} with a slow learning rate), or with the analytical expression for the eigenvalue evolution (Eq.~\eqref{eq:JalphatGaussianEBM_lambert}). In both cases, we utilized empirical covariance matrices extracted from a finite number of samples $M$, sampled from the distribution~\eqref{eq:GEBM_pdf} with  finite $N$. These results are almost insensible to the choice of the population spectrum as discussed in Appendix~\ref{app:different_spectra_and_wasserstrain}.

We demonstrate that the phenomena of overfitting and finite-$M$ corrections can be accurately modeled using RMT to predict the \(N, M \!\to \! \infty\) limit, thereby removing the need for empirical data. Detailed methodologies are provided in Appendix~\ref{sec:RMT}. For a constant \(\rho \!=\! M/N \!=\! 1.5\), Fig.~\ref{fig:rmt_comparison_dynamics}-\figpanel{a} compares the reconstruction error of \(\bm J(t)\) over the training period for various \(N\) values (colored lines) against the asymptotic RMT prediction (dashed black lines), showing strong consistency as \(N\) increases. This agreement extends to the evolution of the test LL (not shown) and the timing of the minimum error and peak test LL as functions of \(\rho\) (see Fig.~\ref{fig:rmt_comparison_dynamics}-\figpanel{b}). Notably, the optimal stopping times for the two estimators do not coincide, yet finite $(N,M)$ trainings align precisely with the asymptotic predictions.

\section{Protocols to mitigate overfitting} 
\begin{figure}[t]
    \centering
\begin{overpic}[width=\columnwidth, trim = 0 30 0 0]{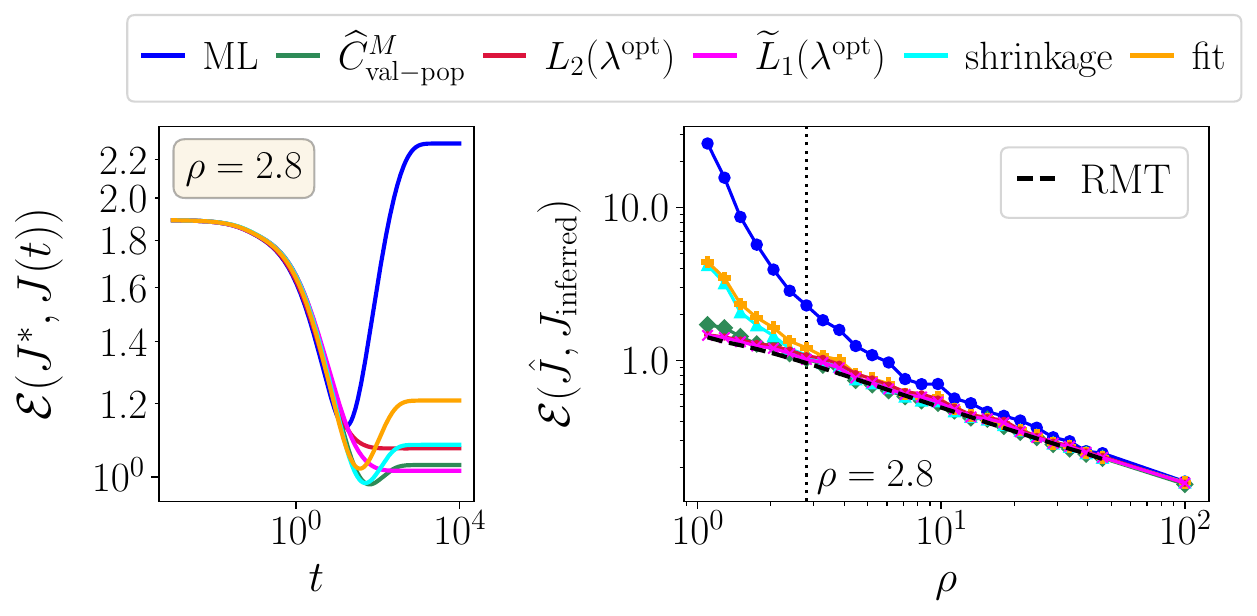}
\put(15,6){{\figpanel{a}}}
\put(68,28.5){{\figpanel{b}}}
\end{overpic}
\caption{Effect of data-correction protocols on the training a GEBM (in \figpanel{a}, for $\rho=2.8$) and on the final model's quality as a function of $\rho$ (in \figpanel{b}): comparison of the reconstruction error $\mathcal{E}_{\text{J}}$ between training from an empirical covariance matrix $\CM$ (blue), optimal $L_2$-regularization (w.r.t. reconstruction, red), shrinkage formula (cyan). The settings are the same as in Fig.\ref{fig:gaussian_model_differentM}.\label{fig:datacorrection} } 
\end{figure}

In the GEBM, non-monotonic behavior stems from adjustments to the eigenvalues of \(\CM\) compared to the population covariance matrix. In fact, one can easily check that replacing the population eigenvectors while retaining \(M\)-dependent eigenvectors to form an optimally corrected matrix, \(\Cvalpop \egaldef \sum_{\alpha} \Cpopval \CMvec {\CMvec}\,\!^{\top}\), almost eliminates the non-monotonic effects on model quality and overfitting, as shown in Fig.~\ref{fig:datacorrection}~\textbf{(a)} (green), and in other  datasets or $\rho$ values we see the bump  completely disappear.
While effective, this approach is useless for real experiments where the population matrix is unknown. Nonetheless, this idealized scenario informs the design of protocols aimed at minimizing overfitting and reducing reliance on uncontrollable early-stopping strategies. We now explore common strategies to mitigate overfitting within our framework, focusing on regularization and shrinkage corrections. We also introduce a versatile downsampling-guided mode-fitting scheme that allows circumvent the traditional limitations of RMT strategies, and design corrections that should be valid beyond GEBMs.

\textbf{Regularization.} In machine learning, regularization priors are standard for preventing overfitting. In the GEBM, they constrain the growth of eigenvalues \(J_{\alpha}\), avoiding suboptimal fixed points affected by mode fluctuations in \(\CM\). For training dynamics, \(L_2\) regularization is applied to the coupling matrix \(\bm J\), and similar outcomes are achieved with projected \(L_1\)-regularization on \(\bm J\)'s eigenbasis, (a protocol that facilitating asymptotic RMT analysis). The impact of regularization at the fixed point is studied for finite \(N\) and in the \(N \to \infty\) limit via RMT. Fig.~\ref{fig:rmt_comparison_dynamics}~\figpanel{c} shows the reconstruction error as a function of \(\lambda\), with empirical results aligning closely with RMT predictions. An optimal \(\lambda_\mathrm{opt}\) minimizes the error but does not match to the value that maximizes the test LL (see \figpanel{d}), complicating \(\lambda_\mathrm{opt}\)'s identification without knowing the  population parameters, akin to identifying optimal early stopping. Further details on regularized training and RMT are provided in Appendices~\ref{subapp:regularization} and \ref{sec:RMT}. Red line in Fig.~\ref{fig:datacorrection}-\figpanel{a} illustrates the error over time for a $L_2$ regularized training using the optimal parameter  \(\lambda^\mathrm{opt}\).

\textbf{Shrinkage correction protocols} are pivotal in statistical learning and signal processing for estimating covariance matrices, particularly when the sample size is small relative to data dimensionality~\cite{bouchaud_cleaning}. Some of these protocols use rotationally invariant estimators (RIEs) to adjust eigenvalues distorted by sampling noise~\cite{ledoitWolf_linear,ledoit_wolf_NL}, while preserving eigenvectors, ensuring corrections are independent of the coordinate system. Based on RMT, RIEs align the eigenvalues of finite-sample covariance matrices to minimize the deviation of the covariance matrix from the population one. Using the optimal RIE from \cite{bouchaud_cleaning}, we correct our $\CM$ matrices, and use them for training our GEBMs. Depicted in light blue in Fig.~\ref{fig:datacorrection}-\figpanel{a}, this new training shows significant improvements in model inference quality, although some non-monotonic behaviors persist. The main drawback of this approach is that it is specific to the GEBM case.

\textbf{Polynomial fit of eigenmodes.} To overcome the limitation of RIEs, we introduce a simple strategy to correct empirically the eigenvalues of $\CM$: the idea is to downsample our dataset to obtain $\widehat{\bm C}^{m}$ with $m<M$ and use the corresponding eigenvalues $\hat{c}_\alpha^m$ values to extrapolate the $m\to\infty$ limit from a linear fit in $1/m$ (as expected from ~\cite{baik2006eigenvalues}). Additional details are given in Appendix~\ref{subapp:fit}. We then use the extrapolated eigenvalues to clean our covariance matrix and run a new training. The evolution of the reconstruction error is shown in Fig.~\ref{fig:datacorrection}-\figpanel{a} (yellow).

\textbf{Comparison of strategies.} The effectiveness of various strategies to counteract overfitting is depicted in Fig.~\ref{fig:datacorrection}-\figpanel{b}, presenting the reconstruction error across different $\rho$ values. Notably, the optimal \(L_2\) regularization, the \(\Cvalpop\) strategy, and the performance at the optimal early stopping point derived from RMT all follow similar trajectories, with a $1 \!/\! \sqrt{\rho}$ scaling for large $\rho$ as expected. While the high performance of these strategies stems from knowing the true model to optimize parameters—not usable in practice—we demonstrate that similar performance can be achieved with RMT-based shrinkage corrections or empirical polynomial fits. These methods do not require prior knowledge of the model, making them especially suitable for real-world inference applications.

\begin{figure}[t!]
\begin{overpic}[width=\columnwidth]{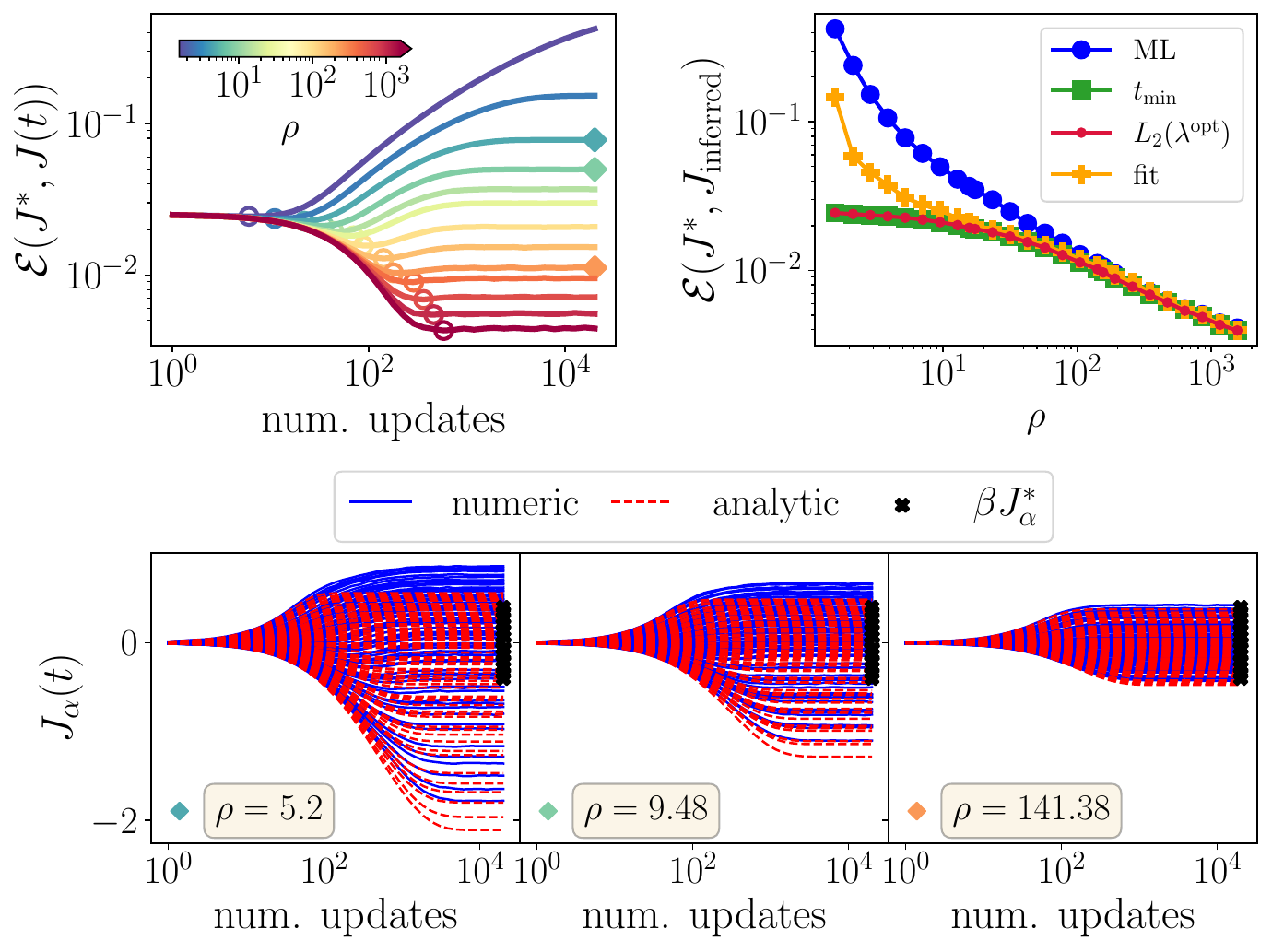}
\put(5,70.5){{\figpanel{a}}}
\put(57,70.5){{\figpanel{c}}}
\put(13.5,27){{\figpanel{b1}}}
\put(42,27){{\figpanel{b2}}}
\put(71,27){{\figpanel{b3}}}
\end{overpic}
\caption{ Results on the BM for the inverse Ising problem. \figpanel{a}-\figpanel{b}: Training dynamics. \figpanel{a}: Reconstruction error (Frobenius norm) vs time (number of updates), for different values of dataset's size. \figpanel{b1}-\figpanel{b2}-\figpanel{b3}: evolution of eigenmodes $J_{\alpha}$ during training for $3$ values of $\rho$: comparison between numerics (blue lines) and analytic curve (red). Black crosses indicate the true models' eigenvalues $\beta \hat{J}_{\alpha}$. \figpanel{c}: effect between data-correction strategies on the inferred model. Comparison between optimal $L_2$-regularization (red), standard ML-training (blue), modes-fitting (orange) and best reconstruction computed at $t_{\mathrm{min}}$ (green).
\label{fig:BM}}
\end{figure}

\section{Boltzmann Machine for inverse Ising\label{sec:BM}}
We extend our analysis to the Boltzmann Machine (BM) or the so-called {\em inverse Ising problem}~\cite{zecchinareviewinverseising}, adapting our approach to binary variables $\bx \!=\! \{\pm 1\}^N$. This model is able to capture multimodal distributions through its pairwise energy function \(E(\bx)\! =\! -\sum_{i<j} J_{ij} x_i x_j \!- \!\sum_i h_i x_i\), with parameters \(\bm \theta\! =\! (\bm J, \bm h)\). Due to the lack of closed-form solutions for the correlation functions and likelihood in BMs, we employ a mean-field approximation suitable e.g. at high temperatures. This approximation allows for an analytic, albeit not exact, expression linking the model's correlation matrix \(\bm C\) to the coupling matrix \(\bm J\) as \(\bm C = (\mathbb{I}_N - \bm J )^{-1}\), facilitating an analytical exploration of the training dynamics shown in \cite{nonconvergentsamplingICML} and further elaborated in Appendix~\ref{app:BMcalculations}.

Similar to GEBMs, an analytical description of spectral dynamics can be applied to BMs, though with certain limitations due to two main factors: a) the ML estimator for the coupling matrix \(\bm J(t)\) may not strictly preserve the same eigendecomposition as \(\CM\), despite typically observing a nice alignment for the strongest modes; and b) the binary nature of the variables is not accounted for in the diagonals of the covariance matrices. We must remove the diagonal constraint to allow independent evolution of the modes, similar to the GEBM scenario, since a spherical constraint as in~\cite{Fanthomme_2022} would introduce mode coupling. This decision is crucial for deriving approximate analytical expressions for training dynamics; without it, the problem becomes intractable in time. However, the proper fixed point alone can still be effectively analyzed using mean field techniques ~\cite{kappenrodriguez_MFBM}.

As detailed in Appendix~\ref{app:BMcalculations}, we can project the gradient on the spectral basis of \(\CM\) and obtain an approximate analytic expression for the evolution of the eigenvalues of \(\bm J\):
\begin{eqnarray}
    &\tau\frac{\der J_\alpha}{\der t} \approx \CMval \! -\! \frac{1}{1 - J_\alpha}, \label{eq:grad_Ja_ising}&\\
    & J_{\alpha} \paren{t} \approx 1 - \frac{1}{\CMval } - \frac{1}{\CMval } W_0 \caja{B_\alpha e^{- \paren{\CMval}^2 \frac{t}{\tau}}},&\label{eq:Jalpha_lambert_Ising}
 \end{eqnarray}
whose fixed point is $J_{\alpha}^{\infty} \!\approx\! 1 \!-\!(\CMval)^{-1}$. This fixed point is shifted due to our unconstrained diagonal, and neither is \(\bm J\) traceless, as compared to the complete treatment. However, Eq.~\eqref{eq:Jalpha_lambert_Ising} still qualitatively captures the training dynamics of BMs, revealing significant differences from the GEBM scenario where \(J_{\alpha}\) and \(c_\alpha\) are no longer inversely proportional. Notably, the smallest \(c_\alpha\) values are associated with negative \(J_\alpha\) values which are not necessarily small in absolute terms, which significantly contribute to the reconstruction of \(\bm J\). Additionally, for positive \(J_\alpha\),  \(J_\alpha\) increases when $\CMval$ does.

\textbf{Results.}
We conducted numerical experiments training an Ising-BM on equilibrium data sampled from a 2D Ising model (i.e. defining $\Jtrue$ on a 2D nearest neighbors lattice) with $N = 8 \times 8$ spins at high temperature (\(\beta = 0.1\)). Figure~\ref{fig:BM} presents the results: Panel \figpanel{a} shows the reconstruction error between the trained model and the ground truth \(\beta \Jtrue \) for different \(\rho\) values, revealing a non-monotonic trend at low \(\rho\) that mirrors observations made with GEBMs (a behavior which is robust w.r.t. the system size, see Fig.~\ref{fig:BM_larger_L32} in Appendix~\ref{app:BMcalculations}). Panels \figpanel{b1}-\figpanel{b3} track the eigenvalue evolution during training for three \(\rho\) values, comparing numerical results (eigenspectrum of \(\bm J(t)\)) with the analytic curve from Eq.~\eqref{eq:Jalpha_lambert_Ising}. While the trends align qualitatively, particularly in capturing the separation of time scales, the analytic curves consistently underestimate the actual eigenvalue evolution due to the overlooked diagonal constraint in the BM model. Nonetheless, this timescale separation is similar to that observed in the GEBM: stronger covariances (\(\CMval > 1 \leftrightarrow J_{\alpha} > 0\)) are learned faster, while weaker covariances (\(\CMval < 1 \leftrightarrow J_{\alpha} < 0\)) take longer. This pattern indicates that the training dynamics are dominated by the convergence of weak $\CMval$. In sparse Ising models, this involves learning the negative spectrum of \(\bm J\). Unlike in GEBMs where negative eigenvalues do not exist, in BMs, these later-encoded modes significantly impact the overall reconstructed \(\bm J\) due to their large absolute eigenvalues, even though they have negligible effect in sampling quality. Accurately inferring sparse Ising models hinges on effectively learning weaker covariances, heavily influenced by finite-data noise. However, training good generative models is considerably faster, see Appendix~\ref{app:BMcalculations}. 

Strategies similar to those used in GEBMs can be employed to mitigate overfitting in BMs, with comparable outcomes as shown in Fig.~\ref{fig:BM}-\figpanel{c}. Shrinkage formulas are not applicable to BMs, yet the empirical polynomial fit correction for the eigenvalues proves still effective at reducing overfitting effects. This is a very good outcome as it is the only non-informative correction (as the identification of $t_\mathrm{min}$ or $\lambda_\mathrm{opt}$ requires knowing $\Jtrue$).
\section{Theoretic extension to generic EBM learning}
The analysis of overfitting for simple models such as the GEBM or the BM constitutes a preparatory  attempt to address this question in the broader context of EBMs. 
Let us see now to which extent the analyses of overfitting carried out so far is also relevant for more general EBMs. 
The line of arguments bears  some similarity with the one justifying that high-dimensional linear regressions are
relevant to  analyze deep learning~\cite{belkin2018understand,hastie2022surprises}.
To this end let us consider the score-matching algorithm~\cite{hyvarinen2005estimation}
as a theoretical proxy for the analyses of overfitting in EBM. Even though this approach might appear sub-optimal in many circumstances  
we postulate that the mechanisms leading to overfitting have similar origin as in more sophisticated methods.
Consider a generic EBM of the form
$p(\x\vert\bm \theta) = Z^{-1} \paren{\bm \theta} e^{-E(\bm x \vert\bm \theta)}$ where $\theta\in\mathbb{R}^P$ is the vector of parameters and having a train set $\D = \{\x_i,i=1,\ldots M\}$ of size $M$.
Defining the score function  as $\psi(\x\vert\theta) \egaldef -\nabla_\x E(\x\vert\theta)$, the score matching loss is given by $\LL_{\rm SM}(\theta) = \frac{1}{2}\hat\E_{\x}\Bigl[\bigl\Vert\bigl(\psi(\x\vert\theta) -\nabla\log\hat p(\x)\bigr\Vert^2\Bigr]$
which, thanks to a by part integration rewrites  as
\begin{equation}
\LL_{\rm SM}(\theta) = \hat\E_{\x}\Bigl[\frac{1}{2}\bigl\Vert\nabla\bigl(E(\x\vert\theta)\bigr\Vert^2 - \Delta E(\x\vert\theta) \Bigr] +{\rm Cst}.    
\end{equation}
Notice first that for the GEBM, this leads to a learning dynamics of the coupling matrix corresponding to $\frac{dJ(t)}{dt} = -\bigl(\hat C J(t)+J(t)\hat C\bigr)+\I$
leading to the solution
\begin{equation}\label{eq:j_Gauss}
j_t(x) = \frac{1-e^{-xt}}{x},
\end{equation}
when assuming that  the initial condition commutes with $\hat C$.
More generally, the dynamics of the score function is governed by a neural tangent kernel (NTK)~\cite{JaGaHo_2018}.
We have
\begin{equation}
\frac{d\psi(\x\vert\theta_t)}{dt} = -\hat\E_{\x'}\Bigl[K_t(\x,\x')\bigl(\psi(\x'\vert\theta_t)-\nabla\log\hat p(\x')\bigr)\Bigr]
\end{equation}
with $K_t(\x,\x') = \partial_{\theta^\top}\psi(\x\vert\theta_t) \partial_{\theta}\psi(\x'\vert\theta_t)^\top$. Integrating by parts the second term we obtain
\begin{equation}\label{eq:NTKdyn}
\frac{d\psi(\x\vert\theta_t)}{dt} = -\hat\E_{\x'}\Bigl[K_t(\x,\x')\bigl(\psi(\x'\vert\theta_t)\Bigr] + \hat \phi_t(\x)
\end{equation}
with $\hat\phi_t(\x) = -\hat\E_{\x'}\Bigl[\partial_{\theta^\top}\psi(\x\vert\theta)\partial_\theta\nabla_{\x'}^\top\psi(\x'\vert\theta)\Bigr]
= -\hat\E_{\x'}\Bigl[\nabla_{\x'}\cdot K_t(\x,\x')\Bigr]$.
As for supervised learning we expect a kernel regime for large enough network's width~\cite{ChOyBa}. 
Then $K$ becomes deterministic, the dynamics is linear, with $\psi(\x,t)$ an explicit function of the kernel matrix $K(\x_s,\x_{s'})$
on the training set. Indeed, the NTK dynamics takes place on a  reproducing kernel Hilbert space (RKHS) of finite dimension corresponding either to ${\mathcal H}_{P} \egaldef {\rm Span}\{\partial_{\theta_q}\psi(\x\vert\theta),q=1,\ldots P\}$,
or to the ${\mathcal H}_{M} \egaldef {\rm Span}\{K(\x,\x_s),s=1,\ldots M\}$, 
depending respectively on whether we are in the under or over-parameterized regime.
In the latter case 
$\hat K_{ss'}\egaldef \frac{1}{M}K(\x_s,\x_{s'})$
is full rank and we have
\begin{equation}\label{eq:dpsi}
\hat\psi(t) = -\frac{1-e^{-\hat K (t-t_0)}}{\hat K}\hat\phi +\hat\psi(t_0)
\end{equation}
where $\hat\psi(t)$ and $\hat\phi$ are respectively the vectors $\{\psi(\x_s\vert\theta_t),s=1,\ldots M\}$ and $\{\hat\phi(\x_s),s=1,\ldots M\}$
and assuming $\hat\psi(t_0)=0$. 
In any case we can consider only the projection of $\psi$ on the RKHS, its transverse part being assumed to be zero at $t=t_0$.
As a result, the dynamics takes place in the ``empirical''  RKHS and we have
\begin{equation}
    \psi(\x\vert\theta_t) = \frac{1}{M}\sum_{s=1}^M K(\x,\x_s) \beta_s(t)
\end{equation}
where the vector $\beta$ is obtained from~(\ref{eq:dpsi}) yielding finally
\begin{equation}
    \psi(\x\vert\theta_t) = -\hat K(\x)^\top \frac{j_t(\hat K)}{\hat K}\hat\phi + \psi(\x\vert\theta_{t_0})
\end{equation}
where $\hat K(\x) = \{\frac{1}{M}K(\x,\x_s),s=1,\ldots M\}$ is the vector of empirical features spanning ${\mathcal H}_E$. Additionally 
$\theta_t$ is directly read off from $\psi(\x\vert\theta_t)$ at first order in $\theta_t$ in the lazy regime
\begin{equation}
\psi(\x\vert\theta_t) \approx \nabla_\theta^\top\psi(\x\vert\theta_{t_0})(\theta_t-\theta_{t_0})
\end{equation}
Using the parameter-sample duality eventually leads to
\begin{equation}
\theta_t = \theta_{t_0} + \frac{j_t(C\n)}{C\n}\phi\n    
\end{equation}
where
\begin{align}
C\n &= \frac{1}{M}\sum_{s=1}^M \nabla_\theta\psi(\x_s\vert\theta)^\top \nabla_{\theta^\top}\psi(\x_s\vert\theta)\\
\phi\n & \egaldef \frac{1}{M}\sum_{s=1}^M \nabla_\theta\psi(\x_s\vert\theta)^\top\hat\phi(\x_s)
\end{align}
In GEBM case we recover~\eqref{eq:j_Gauss} by letting $\psi(\x\vert\theta) = \theta\x$, $K(\x,\x') = \frac{1}{2}(\x\x'^\top+\x^\top\x')$ and  $\hat\phi(\x) = \x $,
leading to $\phi\n = C\n$ with $C\n = \frac{1}{M}\sum_{i=1}^M\x_i\x_i^t$.

\section{Discussion}
This work presents a theoretical framework to understand overfitting in simple energy-based models, using the eigendecomposition of the data covariance matrix to analyze training dynamics. We illustrate how the principal components control a timescale separation, where information progressively encoded from the strongest to the weakest data modes. Due to varying impacts of finite-size noise on different components, this results in an early-stopping point dictated by their interplay. Furthermore, we show that finite sample corrections can be very accurately described analytically using asymptotic RMT analyses. This analysis provide us with an analogous of the GCV in the context of EBM, which deserves further empirical investigations. 
This analysis is exact for Gaussian EBMs and approximate for Ising-BMs at high temperatures, capturing similar phenomena observed in more complex EBMs like RBMs. We discuss data-correction protocols typically used to mitigate overfitting and propose to extend these strategies to more complex models leveraging higher-order data correlations (e.g. by exploiting the SVD decomposition). Further investigations into RMT may clarify how early-stopping points relate to the asymptotic properties of the covariance matrix's spectrum or which should be the proper observables to pinpoint them without a prior knowledge of the data model. Finally, an extension of the theory to EBM via a neural tangent kernel dynamics of the score function deserves further experimental investigations to find relevant hypothesis for the spectrum of  population covariance matrices of tangent features. 

\section*{Acknowledgments} Authors acknowledge financial support by the Comunidad de Madrid and the Complutense University
of Madrid through the Atracción de Talento program (Refs. 2019-T1/TIC-13298 \& Refs. 2023-
5A/TIC-28934), the project PID2021-125506NA-I00 financed by the ``Ministerio de Economía
y Competitividad, Agencia Estatal de Investigación" (MICIU/AEI/10.13039/501100011033), the
Fondo Europeo de Desarrollo Regional (FEDER, UE) and the French ANR grant Scalp
 (ANR-24-
CE23-1320).
\section*{Impact Statement}
This paper aims to advance the field of Machine Learning by deepening our understanding of generative models under data scarcity. While our findings may have broad societal implications, we do not identify any that require specific emphasis at this stage.


\bibliographystyle{icml2025}

\newpage
\appendix
\onecolumn

\section{Derivation of projected gradient equations\label{app:derivation_gradient}}
\begin{figure}[h!]
    \centering
    \includegraphics[width=0.5\textwidth]{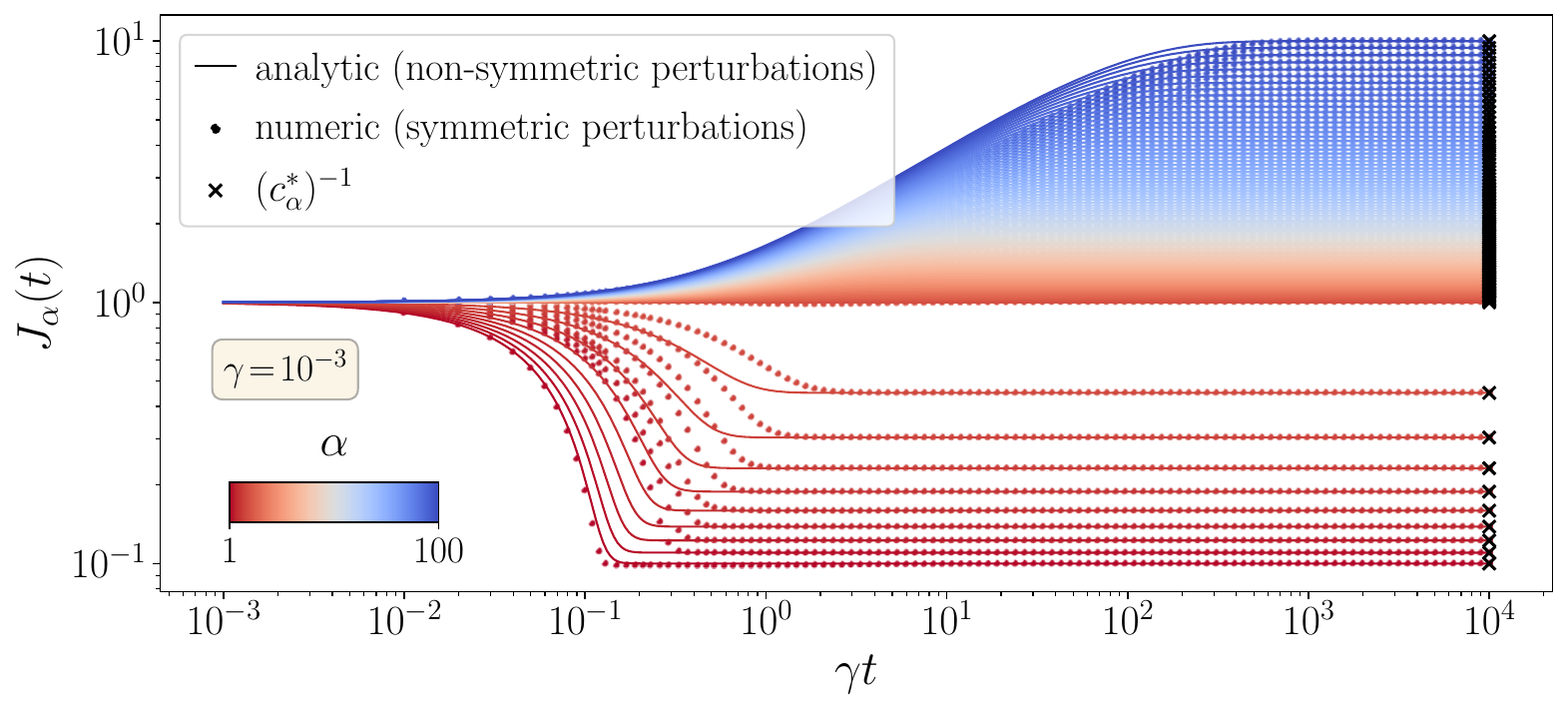}
    \caption{Difference in the eigenvalues' evolution in the training of a GEBM when imposing symmetry or allowing asymmetry in the perturbation of $J_{ij}$. The points correspond to numerical results obtained by enforcing symmetry on $J_{ij}$ after each update during training (i.e. using Eq.~\eqref{eq:gradient_symmetricperturbation}), while the lines represent analytical expressions derived for the case of non-symmetric perturbations (i.e. Eq.~\eqref{eq:JalphatGaussianEBM_lambert}). The setting is the same as Figure~\ref{fig:gauss_training_population} in the main text.
 }  \label{fig:symmetric_perturbation_eigenvalues}
\end{figure}

Starting from the log-likelihood's derivative w.r.t. a parameter $J_{ij}$, we can assume that in the limit of an infinitely small learning rate $\gamma \to 0$ we can replace the discrete-time update equation for the parameter \eqref{eq:gradient_ascent} into a differential equation for the evolution of each parameter $J_{ij}$:
\begin{equation}
     J_{ij}\paren{t+1} = J_{ij}\paren{t} + \gamma \left.\frac{\partial \mathcal{ L}}{\partial J_{ij} }\right|_{\boldsymbol{J}(t)} \longrightarrow  \frac{1}{\gamma} \frac{d J_{ij}}{dt} = \left.\frac{\partial \mathcal{ L}}{\partial J_{ij} }\right|_{\boldsymbol{J}(t)}.
\end{equation}
We now decompose the rhs of the above expression in terms of time-evolution of eigenvalues and eigenvectors of $\bm J$ at time $t$. Given the eigendecomposition $J_{ij}=\sum_\gamma v_i^\gamma J_\gamma v_j^\gamma$, we have
\begin{equation}
    \frac{d J_{ij}}{dt}  = \frac{d}{dt} \sum_\gamma v_i^\gamma J_\gamma v_j^\gamma = \sum_\gamma \left(\frac{dv_i^\gamma}{dt} J_\gamma v_j^\gamma+v_i^\gamma \frac{dJ_\gamma}{dt} v_j^\gamma+v_i^\gamma J_\gamma \frac{dv_j^\gamma}{dt} \right).
\end{equation}
We now project this on the eigenbasis of the eigenvectors of $\boldsymbol{J}$, which after simple algebraic manipulations leads to
\begin{eqnarray}
    \sum_{ij} v_i^\alpha \frac{d J_{ij}}{dt} v_j^\beta & =& \delta_{\alpha \beta } \frac{dJ_\alpha}{dt} + \left(1-\delta_{\alpha \beta}\right)\left(\sum_i v_i^\alpha \frac{d v_i^\beta}{dt} J_\beta + \sum_j \frac{dv_j^\alpha}{dt} J_\alpha v_j^\beta\right) \\&=& \delta_{\alpha \beta } \frac{dJ_\alpha}{dt} + \left(1-\delta_{\alpha \beta}\right)\left( J_\beta- J_\alpha\right) \sum_i v_i^\alpha \frac{d v_i^\beta}{dt} , \label{eq:projected_evolution_JJ}
\end{eqnarray}
where we used the property $\mathrm{d}\left(\boldsymbol{u}^\alpha\cdot \boldsymbol{u}^\beta\right) =0$ because they are vectors of an  orthonormal basis. Finally, combining Eqs.~\eqref{eq:projected_LL_gradient} and~\eqref{eq:projected_evolution_JJ} separating the contributions for $\alpha=\beta$ and $\alpha \neq \beta$ we get to Eq.~\eqref{eq:grad_Ja} in the main text. \\
A final note on the log-likelihoods' gradient: in the first expression ~\eqref{eq:gradient_gaussian_model} we have assumed that perturbations are not symmetric, that is when taking the derivative w.r.t. $J_{ij}$ for $i\neq j$ we assume that $J_{ij}\neq J_{ji}$. Assuming instead \emph{symmetric perturbations} one would get a slight different form of the log-likelihood's gradient w.r.t.~\eqref{eq:gradient_gaussian_model}, given by:
\begin{equation}
 \frac{\partial \like}{\partial J_{ij}}\! = \Lambda_{ij} \caja{- \widehat{C}^M_{ij}\! +\! \left(\bm J^{-1}\right)_{ij}},\label{eq:gradient_symmetricperturbation}
\end{equation}
with $\Lambda_{ij}=1-\delta_{ij} \slash 2$. From the point of view of the training fixed point this is not an issue, the ML estimator is exactly the same in both cases. However, the modified gradient~\eqref{eq:gradient_symmetricperturbation} leads to a slight different dynamics: in particular, it is not anymore true that the dynamics can be exactly decomposed into a separate evolution for the different eigenvalues of $\bm J$ by following the above steps. One could either include symmetric constraints on the eigendecomposition when computing the projected gradient (a more cumbersome process from a mathematical point of view, see e.g.~\cite{magnus99}) or simply double the learning rate on the diagonal terms $i=j$ to compensate for the factor $\Lambda_{ij}$. 
Nevertheless, the difference between the analytic evolution (i.e. Eq.~\eqref{eq:JalphatGaussianEBM_lambert}, obtained assuming non-symmetric perturbation) and a numerical training performed using the gradient~\eqref{eq:gradient_symmetricperturbation} are almost coincident as seen from Figure~\ref{fig:symmetric_perturbation_eigenvalues}. None of the results presented in the manuscript is affected by such a difference in the gradient computation.

\section{Finite-$M$ fluctuations of eigenbasis \label{app:eigenvectors}}
\begin{figure*}[t]
\centering
\begin{overpic}[width=\textwidth,trim=0 30 0 0]{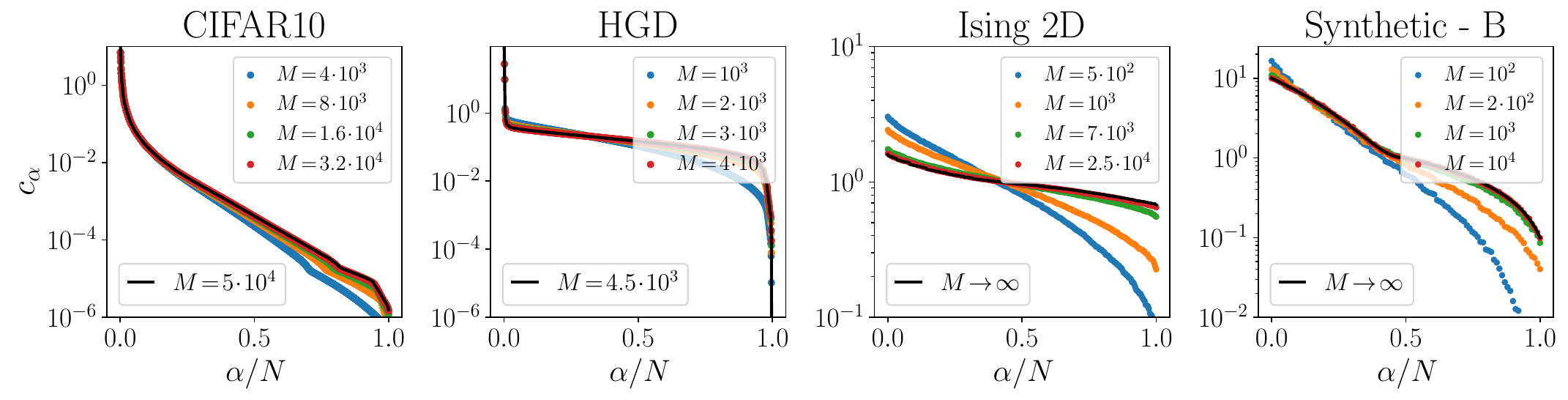}
    \put(5,21){{\figpanel{a}}}
    \put(29,21){{\figpanel{b}}}
    \put(54,21){{\figpanel{c}}}
    \put(78,21){{\figpanel{d}}}
\end{overpic}
\caption{\figpanel{a}-\figpanel{b}-\figpanel{c}: Eigenvalue spectra of the empirical covariance matrix of real datasets, respectively CIFAR-10 (in \figpanel{a}) , Human Genome Dataset (in \figpanel{b}), and a dataset made of equilibirum configurations of a $2$-d Ising model of size $N=16^2$ at $\beta=0.1$ (in \figpanel{c}) . Black lines represent the spectrum computed with the full set of available data (of size $M^*$), while scatter colored points show the result for a subset of data $M<M^*$. \figpanel{d}: the black line shows a synthetic population eigenvalue spectrum generated according to \eqref{def:Cpopmodes}, with $N=100$, and $r=0.5$, $\beta=1.0$, $\gamma=0.5$, $x_1=10^{-1}$, $x_2=10$; colored points display the eigenvalues of the empirical covariance matrix $\CM$ computed by sampling $M$ configurations from a GEBM with $\Jtrue = \Cpop^{-1}$  (from~\eqref{eq:GEBM_pdf}) for different values of $M$.\label{fig:eigenvalues_realandsynthetic_appendix}}
\end{figure*}

 We detail additional eigenvalue spectra of covariance matrices from various datasets in Fig.~\ref{fig:eigenvalues_realandsynthetic_appendix}, complementing those in Fig.~\ref{fig:eigenvalues_realandsynthetic} of the main text. The spectra for CIFAR-10~\cite{krizhevsky2009learning} and the Human Genome Dataset~\cite{10002015global} are displayed in \figpanel{a} and \figpanel{b}, respectively. Panel \figpanel{c} illustrates the empirical covariance matrix from equilibrium configurations sampled from a 2D Ising model with periodic boundaries at high temperature (\(\beta=0.1\)), i.e. in a paramagnetic phase. Panel \figpanel{d} presents another synthetic spectrum, generated through a mixture of power-laws from Eq.~\eqref{def:Cpopmodes}, but using a different set of parameters than those used for Fig.~\ref{fig:eigenvalues_realandsynthetic}-\figpanel{b} (and used for the trainings in Figs.~\ref{fig:gauss_training_population} and ~\ref{fig:gaussian_model_differentM}). This new spectrum was used for the figures involving comparisons with RMT (Figs.~\ref{fig:rmt_comparison_dynamics},~\ref{fig:datacorrection}), for numerical stability issues with the integration of the RMT equations.
 Nonetheless, all the results presented in the main text about the training dynamics of the GEBM can be perfectly reproduced on a wide range of the parameters defining the population eigenvalues, so that the qualitative picture that emerges from our analysis is extremely robust with respect to specific details of the spectrum.

\begin{figure}[h!]
    \centering
    \begin{overpic}[height=0.26\textwidth]{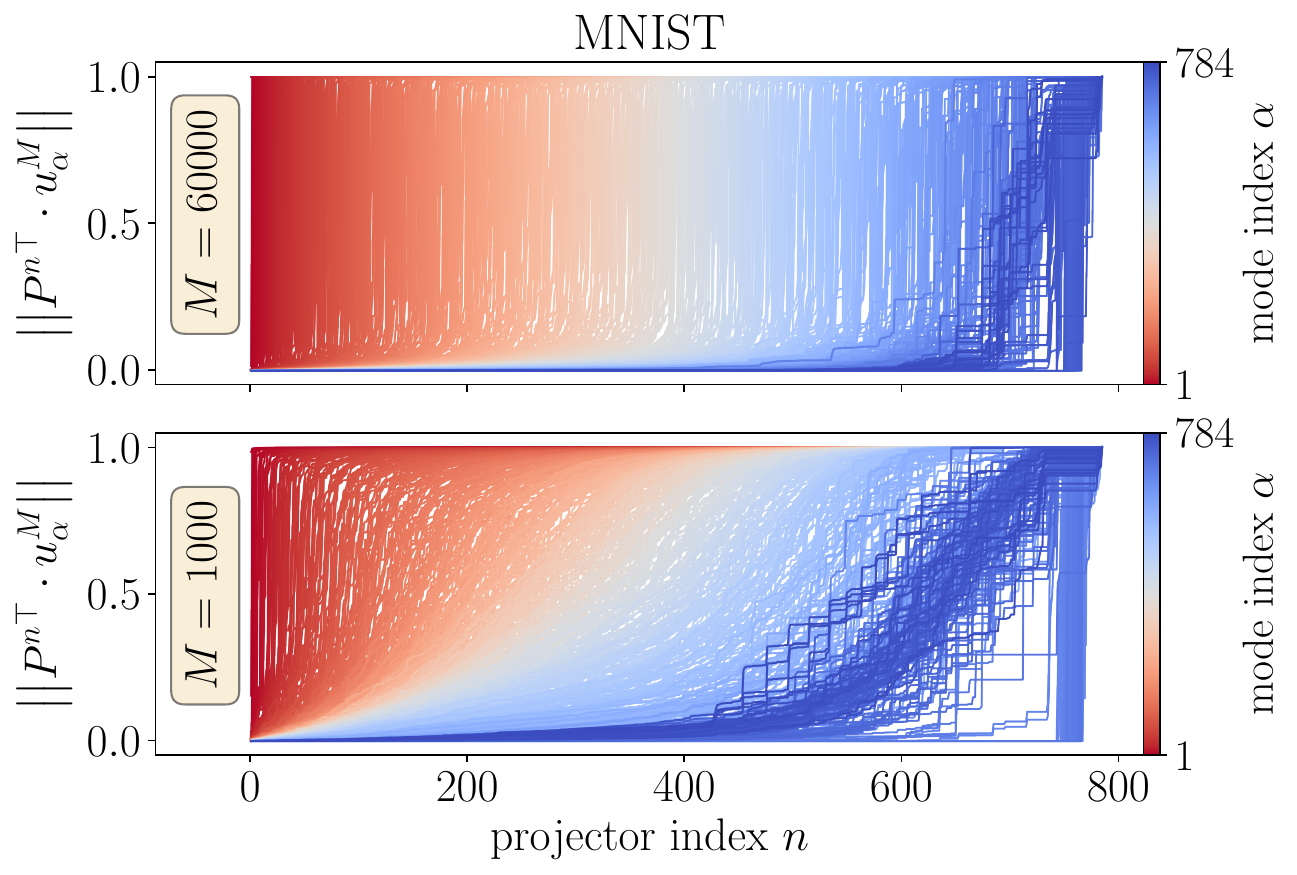}
    \put(1,64){{\figpanel{a}}}
    \end{overpic}\hfill
    \begin{overpic}[height=0.26\textwidth]{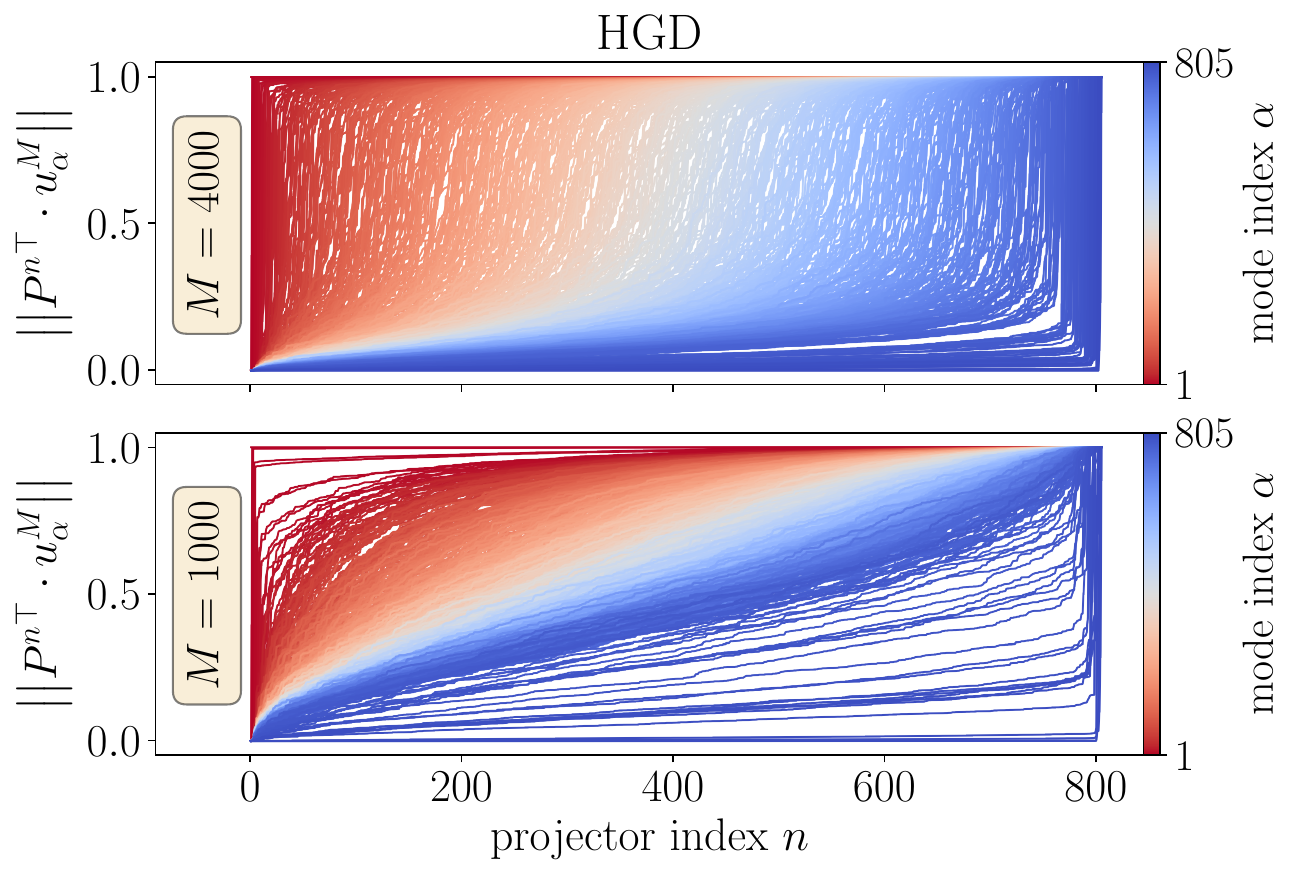}
    \put(1,64){{\figpanel{b}}}
    \end{overpic}
    \vspace{0.5cm}
\begin{overpic}[height=0.26\textwidth]{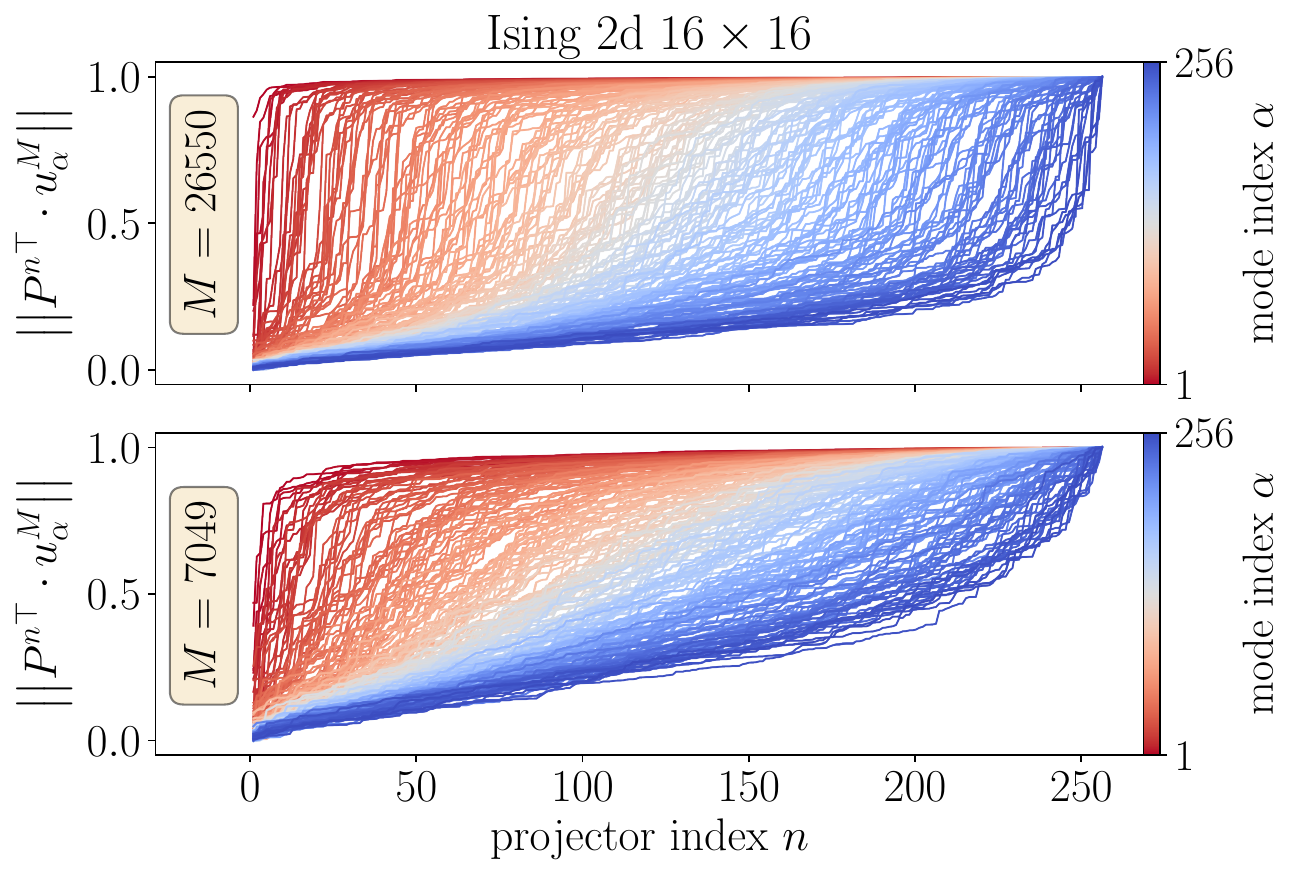}
\put(1,64){{\figpanel{c}}}
    \end{overpic}\hfill
    \begin{overpic}[height=0.26\textwidth]{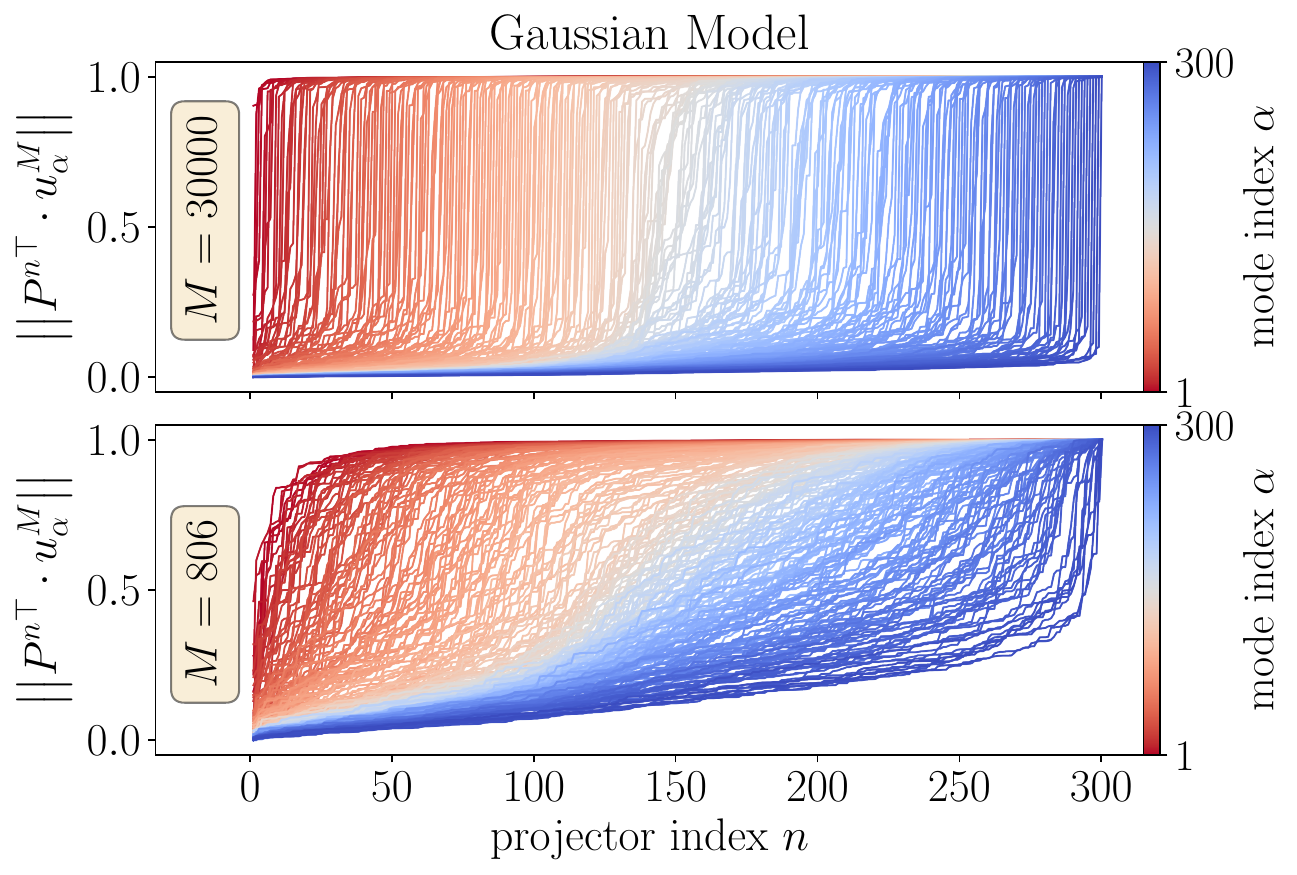}
    \put(1,64){{\figpanel{d}}}
    \end{overpic}
    \caption{Finite-$M$ fluctuations of eigenvectors in the covariance matrix of datasets. The four panels show the norm of the matrix product between the $n$-th projection operator $\bm P^n$, containing the first $n$ eigenvectors of the population matrix $\Cpop$ (for a real dataset, we just take the covariance matrix with the full available data $M^*$) and the $\alpha$-th eigenvector of the covariance matrix $\CM$ with $M<M^*$.  \figpanel{a}-\figpanel{b}-\figpanel{c}  respectively refer to MNIST dataset~\cite{deng2012mnist}, Human Genome dataset, and equilibrium configurations drawn from a 2d Ising model (same setting as in Fig.~\ref{fig:eigenvalues_realandsynthetic_appendix}-\figpanel{c}. \figpanel{d} refers to a synthetic Gaussian Model generated as discussed in the main text with the same settings as in Fig.~\ref{fig:eigenvalues_realandsynthetic}. All panels show the results for two values of $M$, a larger one at the top and a lower one at the bottom. Results are plotted w.r.t. the projector index $n$ and each line correspond to a different $\alpha$. 
    \label{fig:eigenvectors_appendix}}
\end{figure}

 Fig.~\ref{fig:eigenvectors_appendix} illustrates how the eigenbasis of the covariance matrix for real datasets remains consistent against downsampling. Starting with the eigenbasis decomposition of the covariance matrix for the largest available dataset $M^*$—considered our closest approximation to the population matrix $\widehat{\bm C}^{M^*} \approx \widehat{\bm C}^{\infty} = \Cpop$—we denote its eigenvector matrix as ${\bm U}^* = \left\{ {\bm u}^*_{\alpha} \right\}_{\alpha=1}^N$. These eigenvectors are arranged columnwise and sorted in descending order by their corresponding eigenvalue. For each reduced sample size $M<M^*$, we perform a similar decomposition on the resultant empirical covariance matrix $\CM $, with its basis represented as $\bm U^M = \left\{\CMvec \right\}_{\alpha=1}^N$. To evaluate the preservation of eigenvectors, we calculate the norm of the matrix product between a projection operator $\bm P^n$ ---defined as $\bm P^n = \bm U^*_{1:n}$ (incorporating the first $n$ eigenvectors of $\Cpop$)--- and the $\alpha$-th eigenvector of $\CM$, $\CMvec$. This measurement determines whether $\CMvec$ falls within the subspace spanned by the first $n$ eigenvectors of $\Cpop$, thereby helping to mitigate eigenvector oscillations due to exchanges between the ordering of the associated eigenvalues.
 Fig.~\ref{fig:eigenvectors_appendix} shows the norm $\left\lVert {\bm P^n}^\top \cdot \CMvec \right \rVert $ plotted versus $n$ and for each value of $\alpha$, for the same $4$ datasets of Fig.~\ref{fig:eigenvalues_realandsynthetic_appendix}. 
We can observe that for high values of $M$ most eigenvectors are well preserved: this means that most eigenvectors of $\CM$ are contained in the subspace spanned by $\Cpop$, as the norm of such a matrix product raises sharply to $1$ when $n \approx \alpha$. On the other hand, when $M$ is lowered (lower panels on each subplot) the conservation starts to deteriorate, especially in the middle-lower part of the spectrum.  
Interestingly, we observe that the most conserved directions (at least in \figpanel{a}-\figpanel{b}) are both the strongest covariance modes and the lowest ones, a phenomenon already highlighted in \cite{bouchaud_bun_potters_eigenvectors}.

\section{Training dynamics in GEBM with non-commutative initialization\label{app:GEBM_eigenvectors}}
This section provides a brief follow-up to what discussed in the first part of Section~\ref{sec:training}, concerning the training dynamics of a GEBM. For simplicity we focus here only on the infinite-sample scenario (i.e. when training from $\Cpop$), although the same reasoning holds also for finite data. We are also interested in describing the training dynamics for a generic initialization of the matrix $\bm J$, which in general will not commute with $\Cpop$. In this scenario, the model has also to learn the eigenvectors of $\Cpop$.
Fig.~\ref{fig:eigenvalues_realandsynthetic_appendix} shows the evolution of the coupling matrix eigenvalues $J_{\alpha}$ according to Eq.~\eqref{eq:JalphatGaussianEBM_lambert} (shown with solid lines), in comparison to a numerical training done iteratively maximizing the likelihood as in Eq.~\eqref{eq:gradient_ascent} (points). The initial condition here is a matrix $\bm J(0)$ constructed from a random population of modes $J_{\alpha}(0) \sim U[0,1]$ and projected on to a random orthogonal matrix. In this way, $\bm J(0)$ and $\Cpop$ do not commute. At the beginning of the training, there is indeed a discrepancy between theory and simulations, because of the wrong assumption of independence of eigenvalues. Once eigenvectors align, the evolution proceeds independently for each eigenvalue and perfectly follows Eq.~\eqref{eq:JalphatGaussianEBM_lambert}.

\begin{figure}[t]
    \centering
\begin{overpic}[width=0.6\columnwidth, trim = 0 30 0 0]{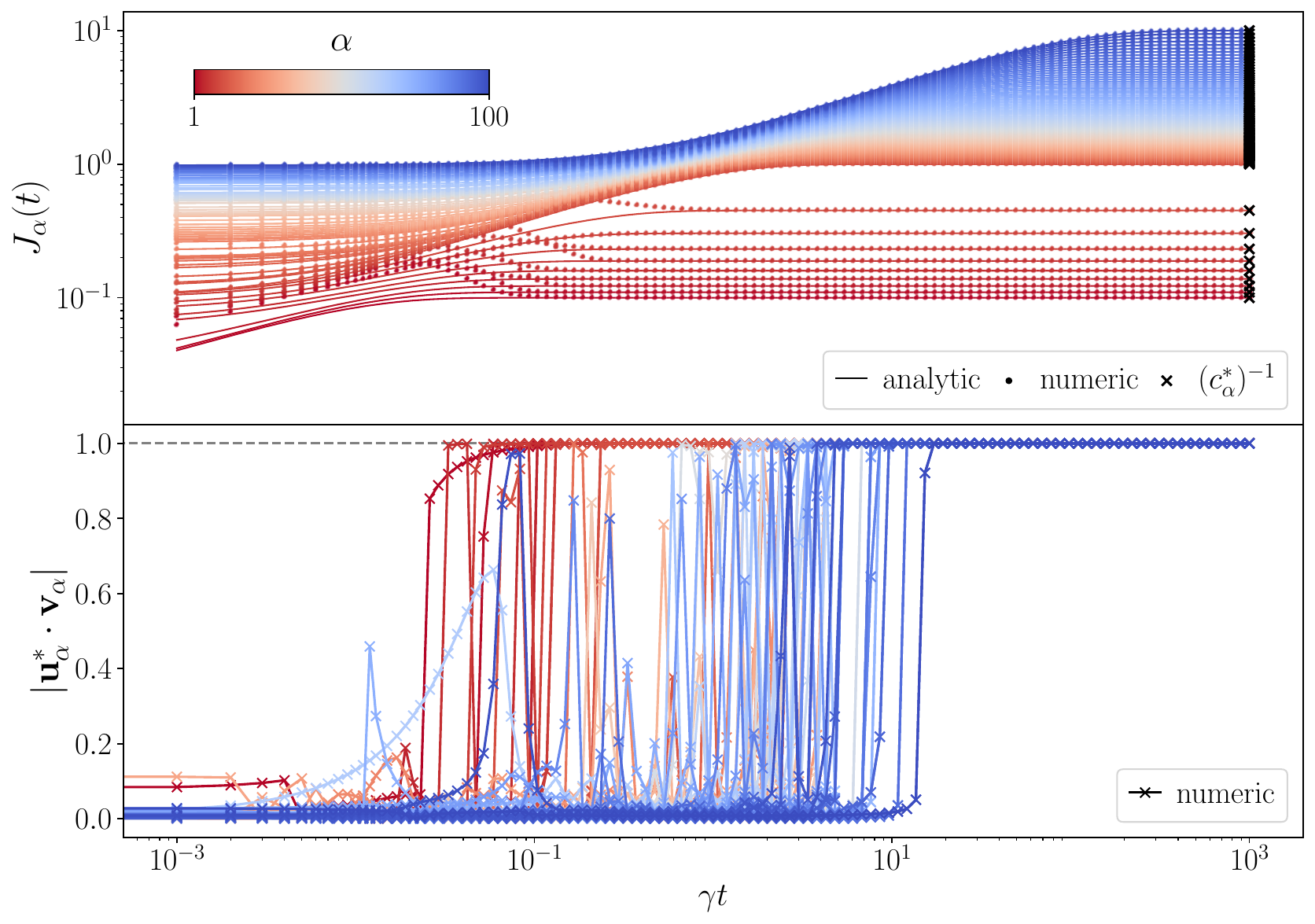}
\put(1,63){{\figpanel{a}}}
\put(1,30){{\figpanel{b}}}
\end{overpic}
\caption{Training dynamics of the GEBM from a population matrix $\Cpop$. The system size and the parameters defining $\Cpop$ are the same as in Fig.~\ref{fig:eigenvalues_realandsynthetic_appendix}-\figpanel{d}. \figpanel{a}: Evolution of eigenvalues, comparison between analytic solution (full line) and numerical training (points). The initial condition $\bm J (0)$ is constructed from a random distribution of modes and projecting it back on to a random orthogonal matrix which differs from the eigenbasis of $\Cpop$, so that the two matrices do not commute. \figpanel{b}: Alignment of eigenvectors, computed as the mode-to-mode overlap between eigenvectors of the population matrix $\Cpopvec$ and eigenvectors of $\bm J$, i.e. $\bm v_{\alpha}$. Red-ish (resp. blue-ish) colors correspond to strong (resp. weak) covariances $\Cpopval$. The learning rate is set to $\gamma=10^{-3}$.\label{fig:GEBM_noncommutative}}
\end{figure}

Note that the initial oscillations of the mode-to-mode eigenvector overlap in Fig.~\ref{fig:GEBM_noncommutative}-\figpanel{b} is due to the fact that eigenvalue learning is non monotonic at the beginning, so that there is an initial exchange in the ordering of the eigenvectors. Nonetheless, after an initial transient all the eigenvectors align to their counterparts in the covariance matrix. This alignment process and with a much faster timescale w.r.t. the learning of eigenvalues (especially the ones associated to weaker covariances):
for this reason, the assumption leading to our analytic description  about independency on the eigenvalues' evolution remains justified for practical purposes. 
This reasoning about eigenvector alignment holds for any input covariance matrix: what determines the non trivial dynamics of the reconstruction error (explained in Sec.~\ref{sec:training}) is fully determined by the noise in the eigenvalues.
\section{Robustness of results w.r.t. initialization at finite $M$\label{sec:other_inits}}
\begin{figure}[h!]
    \centering
\begin{overpic}[width=\textwidth, trim = 0 30 0 0]{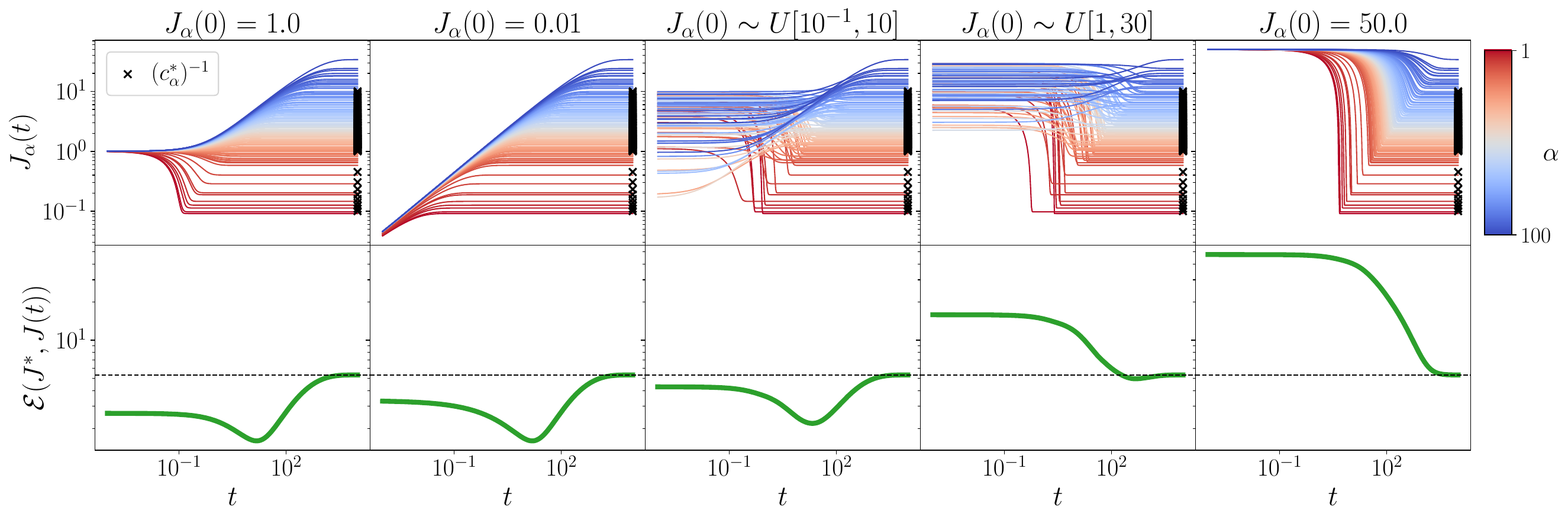}
\put(6.5,16){{\figpanel{a1}}}
\put(24,26){{\figpanel{a2}}}
\put(42,26){{\figpanel{a3}}}
\put(59,16){{\figpanel{a4}}}
\put(77,16){{\figpanel{a5}}}
\put(6.5,13){{\figpanel{b1}}}
\put(24,13){{\figpanel{b2}}}
\put(42,13){{\figpanel{b3}}}
\put(59,13){{\figpanel{b4}}}
\put(77,3){{\figpanel{b5}}}
\end{overpic}
\caption{Results on the training dynamics GEBM at finite amount of data by varying the initial conditions.  Panels \figpanel{a}'s (top row) show  the eigenvalues' evolution (according to Eq.~\eqref{eq:Jalpha_lambert_Ising}), while panels \figpanel{b}'s (bottom row) show the corresponding reconstruction error $\mathcal{E}_{\text{J}}$ w.r.t the ground truth model $\Jtrue$; all quantities are shown versus time. Each column corresponds instead to a different initial condition. From left to right: an identity-like initialization ($J_{\alpha}(0)=1$) in column \figpanel{1}, as in Fig.~\ref{fig:gauss_training_population}-\figpanel{b}; a small-coupling initialization in ($J_{\alpha}(0)=10^{-2}$), in column \figpanel{2}; two random initialization of modes (resp. in the boundaries $[10^{-1}; 10]$ and $[1; 30]$ in column \figpanel{3}-\figpanel{4}); a constant initialization to very high values larger than the fixed point, i.e. ($J_{\alpha}(0)=50 > 1 / \CMval$) in column \figpanel{5}.  All trainings are performed analytically, with an empirical covariance matrix $\CM$ generated with the same settings as in Fig.~\ref{fig:gauss_training_population} with $\rho=M\slash N =2.11$. \label{fig:other_inits}}
\end{figure}

We show in Fig.~\ref{fig:other_inits} some analogous results w.r.t. Fig.~\ref{fig:gauss_training_population}-\figpanel{b} for the training dynamics of the GEBM in the case of finite $M$ (here we set $\rho=M \slash N = 2.11$) by varying the initialization. In this case, we do not care about eigenvectors' alignment as in the previous section: we only consider different initializations for the eigenmodes of the coupling matrix, i.e. $\{J_{\alpha}\}_{\alpha=1}^N$. We can observe how the non-monotonic behavior of the reconstruction error (plotted in the bottom rows for each initialization) is robust against different standard, uninformative and small initializations, and it keeps appearing as long as $J_{\alpha}(0)<1 / \CMval$ for the majority of the eigenvalues (especially the ones corresponding to weak covariances). We can also observe how, increasing the initial conditions to higher values than the fixed point (i.e. moving from columns \figpanel{1}-\figpanel{2}-\figpanel{3} to the right most ones \figpanel{4}-\figpanel{5}) the non-monotonic behavior disappears, indicating that the early-stopping break-point no longer exists and that there is now a way to mitigate the overfitting effects with this strategy.
 However, it is common practice to start with small values. Moreover, in more complex EBMs where sampling is required to estimate the correlations of the model in the LL gradient (e.g., BMs or RBMs), it may be a very bad idea to assume extreme initializations (i.e. far from an uninformative initialization where the parameters of the model are small): this could lead to ergodicity problems in sampling, as the model may get stuck in spin-glass-like phases, a phenomenon that has been well studied in several EBMs (see e.g. \cite{decelle2018thermodynamics}). 
\newpage

\section{Additional generation quality metrics}
In this section we consider  an additional metric to compute the discrepancy between the trained model and the true one, namely the Wasserstrein distance~\citep{Delon_Desolneux_Salmona_2022}.
Figure~\ref{fig:datapaper} shows the same data as in Figure~\ref{fig:gaussian_model_differentM} of the main text, now including the evolution of the Wasserstrein distance between the trained model and the true one w.r.t. training time. Also this quantity shows a non-monotonic behavior in $t$ especially for low $M$, with a clear early-stopping point. In the rightmost panel, we compare the locations of the minima of each error estimator (and the maximum of the log-likelihood) as functions of $\rho$. We observe that the time point corresponding to the minimum Wasserstein distance follows a trend very similar to that of the maximum log-likelihood.

\section{GEBM analysis with various eigenvalue spectra \label{app:different_spectra_and_wasserstrain}}

This section presents additional results on the GEBM, analogous to Fig.~\ref{fig:gaussian_model_differentM} in the main text, obtained using alternative spectra for the population covariance matrix. To assess the robustness of our findings with respect to spectral choice, we replicate the analysis using both synthetic and empirical spectra.

First, in Fig.~\ref{fig:dataold}, we consider a synthetic spectrum distinct from Eq.~\eqref{def:Cpopmodes}: for $N = 100$, we generate 10 dominant modes with amplitudes uniformly distributed in $[2, 10]$, and a bulk of $N - k = 90$ noisy modes with amplitudes in $[10^{-1}, 1]$. The qualitative behavior, including overfitting effects, remains unchanged.

\begin{figure}
\begin{overpic}[width=\textwidth]{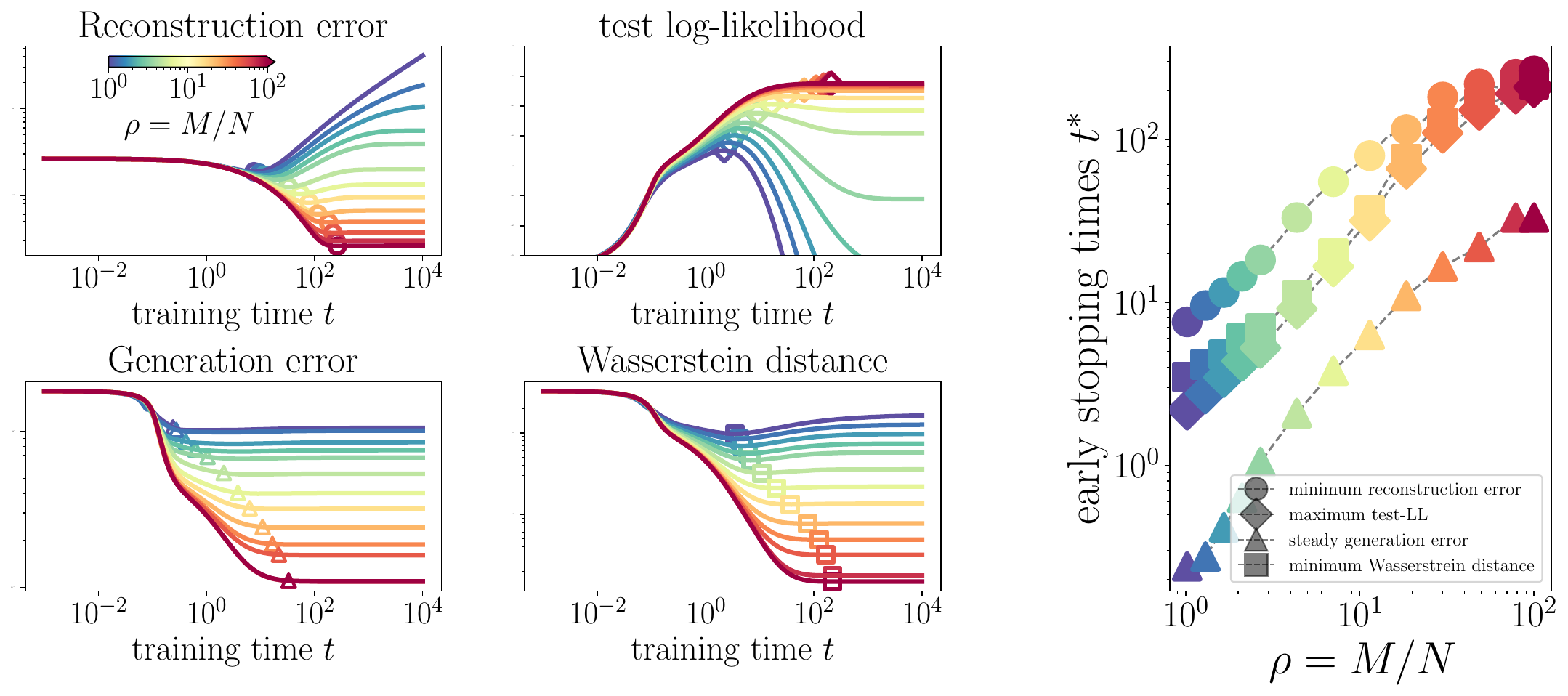}
\put(1,42.5){{\figpanel{a}}}
\put(33,42.5){{\figpanel{b}}}
\put(1,21){{\figpanel{c}}}
\put(33,21){{\figpanel{d}}}
\put(72,43){{\figpanel{e}}}
\end{overpic}    
    \caption{We compare the results shown in Fig.~\ref{fig:gaussian_model_differentM}, obtained using various generation quality measures, with the corresponding curves computed using the Wasserstein distance.
    \figpanel{a}-\figpanel{b}-\figpanel{c}-\figpanel{d} display respectively the reconstruction error \(\mathcal{E}_\text{J}\), the test-LL, the Wasserstein distance and the generation error $\mathcal{E}_\text{C}$, all plotted vs time, for various sample sizes \(M\) (indicated by a color gradient from blue to red for increasing $\rho\!= \!M \!\slash\! N$. Dashed black lines refer to a training from \(\Cpop\) (i.e. \(M \!\to \!\infty\)).  \figpanel{e}: comparison between time of minimum reconstruction (circles), maximum test LL (diamonds), minimum Wasserstein distance (squares) and time at which the generation error converges to its steady-state value. These quantities are also shown in the related panels for better clarity. Apart on panel \figpanel{d}, this figure contains the same information and quantities as Fig.~\ref{fig:gaussian_model_differentM}.}
    \label{fig:datapaper}
\end{figure}

\begin{figure}
    \begin{overpic}[width=\textwidth]{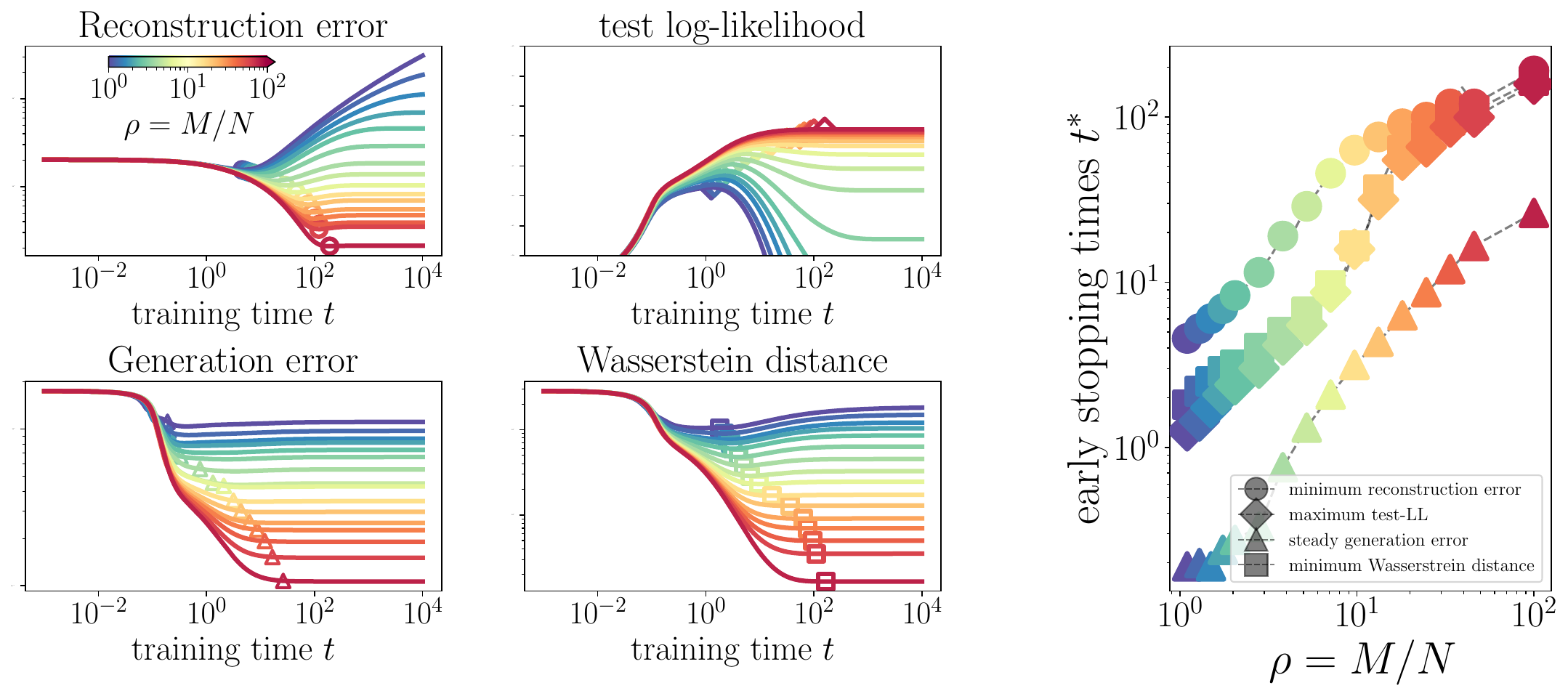}
\put(1,42.5){{\figpanel{a}}}
\put(33,42.5){{\figpanel{b}}}
\put(1,21){{\figpanel{c}}}
\put(33,21){{\figpanel{d}}}
\put(72,43){{\figpanel{e}}}
\end{overpic}    
    \caption{Same plots as in Fig.~\ref{fig:datapaper}, this time obtained by training a GEBM starting from a synthetic population covariance matrix spectrum of dimension $N = 100$, with a set of $10$ dominant modes with amplitudes uniformly distributed in the interval $[2, 10]$, and a bulk of $N - k = 90$ noisy modes with amplitudes uniformly distributed between $10^{-1}$ and $1$.  }
    \label{fig:dataold}
\end{figure}

\begin{figure}
    \begin{overpic}[width=\textwidth]{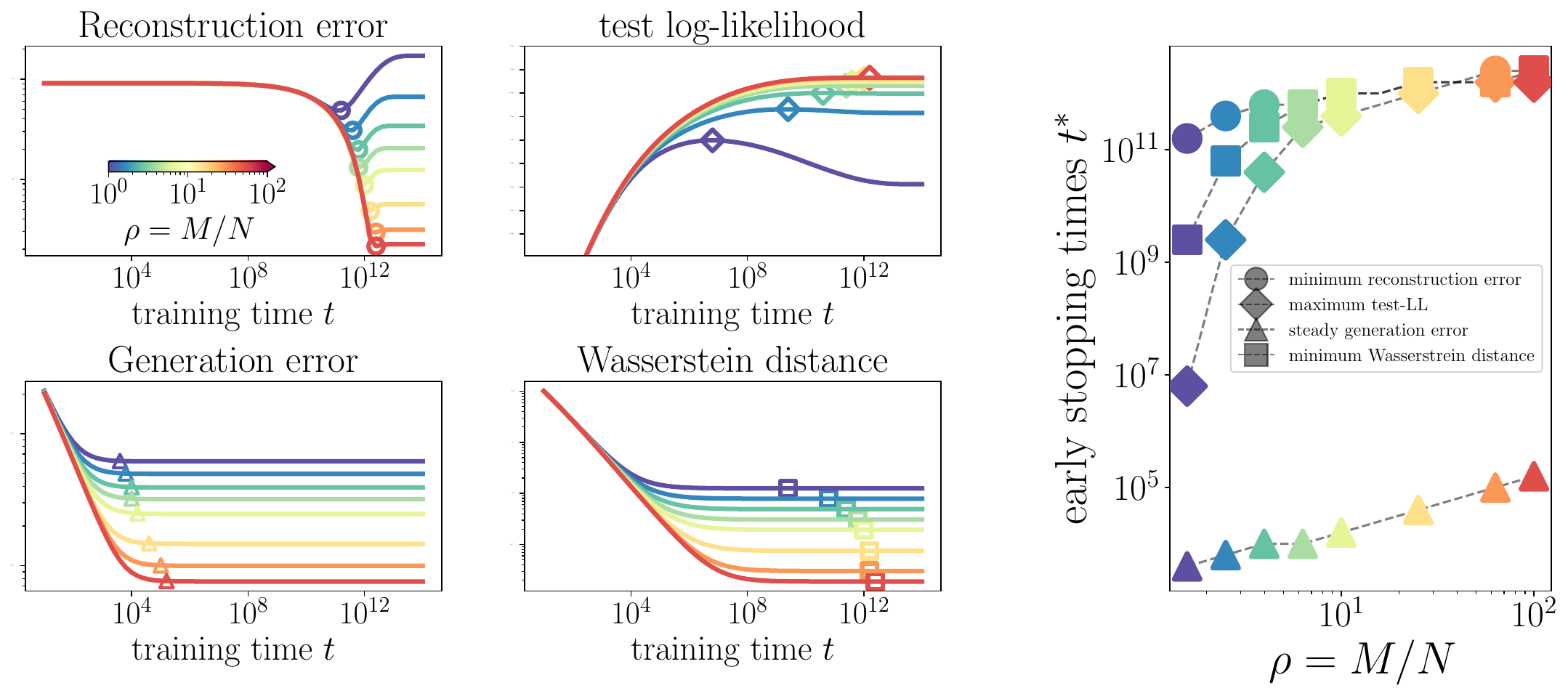}
\put(1,42.5){{\figpanel{a}}}
\put(33,42.5){{\figpanel{b}}}
\put(1,21){{\figpanel{c}}}
\put(33,21){{\figpanel{d}}}
\put(72,43){{\figpanel{e}}}
\end{overpic}
    \caption{Same plots as in Fig.~\ref{fig:datapaper}, this time obtained by training a GEBM starting from the eigenvalue spectrum of the empirical covariance matrix computed from the MNIST dataset, with a cutoff at $10^{-6}$ to filter out weak modes.
    }
    \label{fig:mnist}
\end{figure}

\begin{figure}
\begin{overpic}[width=\textwidth]{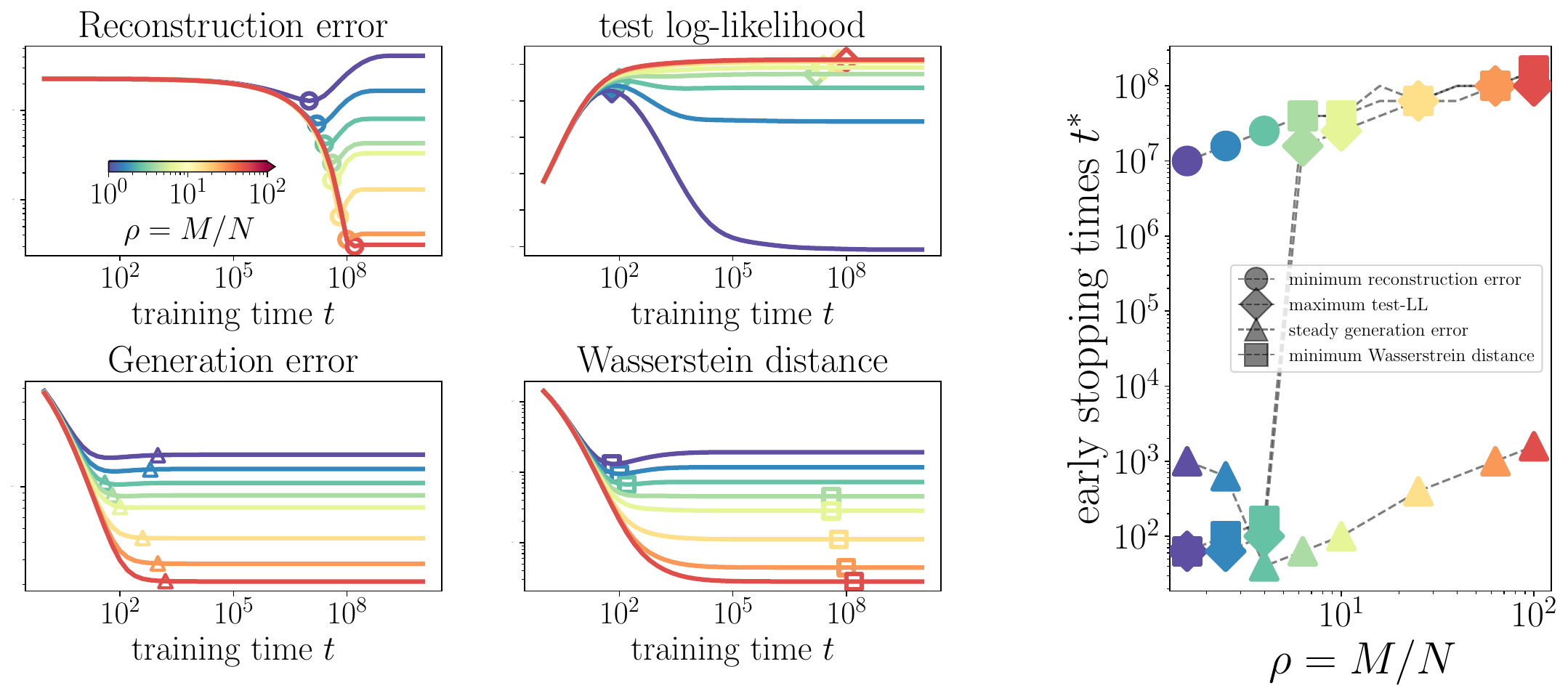}
\put(1,42.5){{\figpanel{a}}}
\put(33,42.5){{\figpanel{b}}}
\put(1,21){{\figpanel{c}}}
\put(33,21){{\figpanel{d}}}
\put(72,43){{\figpanel{e}}}
\end{overpic}
   \caption{Same plots as in Fig.~\ref{fig:datapaper}, this time obtained by training a GEBM starting from the eigenvalue spectrum of the empirical covariance matrix computed from the HGD dataset, with a cutoff at $10^{-12}$ to filter out weak modes.
    }
    \label{fig:hgd}
\end{figure}

Next, we repeat the analysis using empirical spectra: specifically, the eigenvalues of the sample covariance matrices from the MNIST and Human Genome Dataset (HGD), shown in Figs.~\ref{fig:mnist} and \ref{fig:hgd}, are used as population spectra. The resulting dynamics, analogous to Fig.~\ref{fig:datapaper}, again show no qualitative deviations. In all cases, the non-monotonic temporal behavior of key metrics and the early-stopping times are preserved.


\newpage

\section{Asymptotic analysis through Random Matrix Theory}\label{sec:RMT}
\subsection{General case}
Various quantities appearing in the core of the manuscript are explicit function of the empirical covariance matrix $\CM$ and as such are amenable to asymptotic analysis thanks to random matrix theory (RMT).
These quantities are respectively the train, test energy (associated to the EBM), the coupling error and the LL (train and test). For the sake of clarity, we repeat them here:
\begin{align}
E_{\rm train} &= \frac{1}{N}\Tr[\J \CM] ,\label{def:E_train}\\[0.2cm] 
E_{\rm test} &= \frac{1}{N}\Tr[\J \Cpop],   \label{def:E_test}\\[0.2cm]
{\mathcal E}_{\rm J} &\egaldef\frac{1}{N}\Vert \J-\Jtrue\Vert_F^2 \label{def:E_couplings}\\[0.2cm]
LL_{\rm train,test} &\egaldef \frac{1}{2N}\log\det[\J]-\frac{1}{2}E_{\rm train,test},\label{def:LL}
\end{align}
where $\Cpop \egaldef \lim_{M\to\infty} \CM$ 
is the {\em population matrix}, $\Vert\cdot\Vert_F^2$ the Frobenius norm while 
$\J$ is the estimation of the coupling matrix from the train samples $\x$ assumed to be of the form 
$\x = \bm F\z$, with ${\mathbb E}( \bm z \bm z ^\top) = \I$, $\bm F \bm F^\top = \CM $, $\tau = \Vert \x\Vert$ 
distributed w.r.t. some density $\sigma(\tau)$. 
Depending on the setting (dynamical, spectral $\widetilde{L}_1$ or $L_2$) $\J$ may appear in three different explicit functional form $j_t$, $j_\alpha^{(\widetilde{L}_1)}$ and $j_\alpha^{(\rm L_2)}$ of $\CM$. We have
\begin{align}
    j_t(x) &= \frac{1}{x}\paren{1+W_0[-e^{-x^2 t-1}]},\qquad \text{training dynamics} ,
    \label{eq:jt}\\[0.2cm]
    j_\alpha^{(\widetilde{L}_1)}, &=                       
\frac{\alpha}{1+\alpha x},\qquad\qquad\qquad (L_1\ \text{(spectral) regularization}), 
\label{eq:j_L1} \\[0.2cm]
    j_\alpha^{(\rm L_2)}, &=
\frac{\alpha}{2}\paren{\sqrt{x^2+\frac{4}{\alpha}}-x},\qquad\qquad (L_2\ \text{ regularization}).
\label{eq:j_L2}
\end{align}
A derivation of Eqs.~\eqref{eq:j_L1}-\eqref{eq:j_L2} is given in Appendix~\ref{subapp:regularization}. The $j_t$ corresponds to the situation where all eigenvalues $J_\alpha$ have the initial condition
$J_\alpha(0)= 0$ and follow the time evolution of Eq.~\eqref{eq:JalphatGaussianEBM_lambert}. 
Let us call generically $j$ the functions given above. 

Using the resolvant
\[
\G\n(z) \egaldef \frac{1}{z\I-\CM}ç,
\]
we can express the various quantities of interest with help of Cauchy integrals 
\begin{align*}
E_{\rm train} &= \frac{1}{2i\pi}\oint_{\C} dz j(z) \Tr\bigl[\G\n(z)\bC\n\bigr] , \\[0.2cm]
E_{\rm test} &= \frac{1}{2i\pi}\oint_{\C} dz j(z) \Tr\bigl[\G\n(z)\Cpop\bigr],  \\[0.2cm]
{\mathcal E}_{\rm J} &= \Tr\bigl[\Cpop^{-2}\bigr]+\frac{1}{2i\pi}\oint_{\C} dz \Bigl(j^2(z)\Tr\bigl[\G\n(z)\bigr]-2j(z)\Tr\bigl[\G\n(z)\Cpop^{-1}\bigr]\Bigr),\\[0.2cm]
LL_{\rm train,test} &= \frac{1}{2i\pi}\oint_{\C} dz \frac{\log[j(z)]}{2} \Tr\bigl[\G\n(z)\bigr]
-\frac{E_{\rm train,test}}{2},
\end{align*}
where $\C$ is a contour of integration around the real axis.
Next, from RMT, in the proportional asymptotic limit $M,N\to\infty$ with fixed $M/N = \rho$,
$\G\n(z)$ has a deterministic equivalent~\cite{hachem2007deterministic} $\G$, 
defined as
\[
\G(z) = \frac{1}{z\I  - \Lambda(z) \Cpop},
\]
with $\Cpop$ the population matrix and $\Lambda(z)$ is given implicitly by the following self-consistent equations~\cite{MaPa}
\begin{align}
  \Lambda(z) &= \int\frac{\sigma(\tau)}{1-\Gamma(z)\tau},\label{eq:Lbdz}   \\[0.2cm]
  \Gamma(z) &= \frac{1}{\rho}\int \frac{\nu(dx) x}{z-\Lambda(z) x}, \label{eq:Gz}
\end{align}
with $\nu(dx)$ the spectral density of the population matrix and where 
\begin{equation}\label{eq:Gamma}
\Gamma \egaldef \lim_{N,M\to\infty\atop \frac{M}{N}=\rho} \frac{\alpha}{M}\Tr\bigl[\G\n \Cpop].
\end{equation}
For sake of clarity we do not consider the Marchenko-Pastur equations in 
full-generality and actually assume the fluctuation of $\z$ to be negligible i.e. we take 
$\sigma(\tau) = \delta(\tau-1)$.
Letting $\bar\nu(dx)$ the asymptotic limit of the empirical spectrum in the proportional regime, 
its Stieltjes transform is given by the trace of the resolvent:
\[
g(z) \egaldef \int \frac{\bar\nu(dx)}{z-x}.
\]
Then the bulk spectrum is given by the Stieltjes transform
\[
g(y+i\epsilon) = g_r(y)+i\pi\frac{\epsilon}{\vert\epsilon\vert}\bar\nu(y),
\]
which rewrites (disregarding the pole at $z=0$ for $\rho<1$)
\begin{equation}\label{eq:bnu}
\bar\nu(y) = \frac{\rho \Lambda_i(y)}{\pi y} = \frac{\rho}{\pi y} \frac{\Gamma_i(y)}{\bigl[1-\Gamma_r(y)\bigr]^2+\Gamma_i(y)^2}.
\end{equation}
Along the contour we integrate over $z=y+i\epsilon$ with $\epsilon$ infinitesimal.
In the limit $\epsilon\to 0$, both $\Lambda$ and $\Gamma$ may acquire a finite imaginary part
which we write as
\begin{align*}
    \lim_{\epsilon\to 0^\pm}\Lambda(z) = \Lambda_r(y)\pm\Lambda_i(y) \\[0.2cm]
    \lim_{\epsilon\to 0^\pm}\Gamma(z) = \Gamma_r(y)\pm\Gamma_i(y).
\end{align*}
In terms of these quantities we obtain the following equations for the train and test energies:
\begin{align*}
E_{\rm train} &= \frac{\rho}{\pi}\int_0^\infty dy j(y) \bigl[\Lambda_r(y)\Gamma_i(y)+\Lambda_i(y)\Gamma_r(y)], \\[0.2cm]
E_{\rm test} &= \frac{\rho}{\pi}\int_0^\infty dy j(y) \Gamma_i(y)+\ind{\rho<1}j(0) c(\rho),
\end{align*}
where $c(\rho)$ given implicitly by
\begin{equation}\label{eq:c_rho}
\int \frac{\nu(dx)x}{x+c(\rho)} = \rho.
\end{equation}
$E_{\rm train}$ may also be written
\[
E_{\rm train} = \int_0^\infty dy yj(y)\bar\nu(y),
\]
with $\bar\nu(y)$ given in~(\ref{eq:bnu}).

The coupling error takes the form for any $\rho>0$
\begin{align*}
{\mathcal E}_{\rm J} &= \int_0^\infty\frac{\nu(dx)}{x^2}+\int_0^\infty \bar\nu(dy)j^2(y) -\frac{2}{\rho}\int_0^\infty \frac{\bar\nu(dy)}{y}\Bigl[(1-\rho)+2\rho\Lambda_r(y)\Bigr]j(y)\\[0.2cm]
&+\ind{\rho<1}\Bigl[(1-\rho)j^2(0)+2j(0)\Bigl(\frac{1-\rho}{c(\rho)}-\int_0^\infty\frac{\nu(dx)}{x}\Bigr)\Bigr],
\end{align*}
but in practice we consider only the under-parameterized regime corresponding to $\rho>1$. 

\subsection{Special case of spectral $L_1$ Regularization}
The case corresponding to the form~(\ref{eq:j_L1}) can be treated more directly without use of 
Cauchy integrals. In that case, considering instead the resolvant 
\[
\G\n = \frac{1}{\I+ \alpha\CM},
\]
with the inverse penalty 
\[
\alpha = \frac{1}{\lambda}
\]
introduced here for convenience. This 
leads to the following form of the various quantities of interest 
\begin{align}
E_{\rm train} &= \frac{\alpha}{N}\Tr[\G\n \CM] \label{def:E_train}\\[0.2cm] 
E_{\rm test} &= \frac{\alpha}{N}\Tr[\G\n \Cpop ]. \label{def:E_test}  \\[0.2cm]
E_{\rm couplings} &\egaldef\frac{1}{N}\Vert \alpha \G\n-\J\Vert_F^2 \label{def:E_couplings}\\[0.2cm]
LL_{\rm train,test} &= \frac{1}{2N} \Tr[\log \alpha \G\n]-\frac{1}{2}E_{\rm train,test}.
\end{align}
In the scaling limit we again have a deterministic equivalent~\cite{hachem2007deterministic} of the resolvent of the form
\[
\G = \frac{1}{\I+\Lambda \Cpop}
\]
where the fixed point equations now read ($\sigma(\tau) = \delta(\tau-1)$)
\begin{align}
  \Gamma &= \frac{\alpha}{\rho}\int  \nu(dx)\frac{x}{1+\Lambda x} \label{eq:rmt_gamma}\\[0.2cm]
  \Lambda &= \frac{\alpha}{1+\Gamma }      \label{eq:rmt_lambda}
\end{align}
with again $\Gamma$ given by~(\ref{eq:Gamma}).
The expression for $E_{\rm train,test}$ are straightforward in the scaling limit:
\begin{align*}
E_{\rm train} &= 1-\int \frac{\nu(dx)}{1+\Lambda x} \\[0.2cm]
E_{\rm test}  &= \frac{\Gamma}{\rho}.
\end{align*}
Remarkably, thanks to a leave-one out argument there is a deterministic relationship between 
the train and test energy. 
For $s\in{\mathcal I}_{\rm train}$, we have a leave-one out relation of the form
\[
\G\n\x_s = \frac{\G_{\backslash s}\n\x_s}{1+\frac{\alpha}{M}\x_s^t\G_{\backslash s}\n\x_s}
\]
where $\G_{\backslash s}\n$ is the resolvant obtained after removing $s$ from the train set.
Assuming the samples to be of the form $\x_s = F\z_s$ with $FF^t = C$
with  $\z_s = \mathcal{N}(0,M\I)$ such that $\E(\z_s\z_s^t) = \I$, then for large $N$ we have the concentration property 
\[
\x_s^t\G_{\backslash s}\n\x_s = \frac{1}{M}\Tr\bigl[\G\n C \bigr] +\mathcal{O}\bigl(\frac{1}{\sqrt{M}}\bigr).
\]
As a result for large $N,M$ we immediately obtain
\[
E_{\rm train} = \frac{E_{\rm test}}{1+\frac{1}{\rho}E_{\rm test}},
\]
which can be reverted as 
\begin{equation}\label{eq:train-test}
E_{\rm test} = \frac{E_{\rm train}}{1-\frac{1}{\rho}E_{\rm train}}.    
\end{equation}
Concerning the error on the couplings, we have 
\begin{align*}
  E_{\rm J}   &= \frac{1}{N}\Tr\bigl[\bigl(\alpha \G\n-\Jtrue\bigr)^2\bigr]  \\[0.2cm]
  &= \frac{1}{N}\Bigl(\alpha^2\Tr\bigl[{\G\n}^2]+\Tr\bigl[\Jtrue^2\bigr]-2\alpha\Tr\bigl[\G\n \Jtrue\bigr]\Bigr)
\end{align*}
From Ledoit-P\'echet the last term simply reads (up to $\mathcal{O}\bigl(1/\sqrt{M}\bigr)$ corrections):
\[
  \frac{1}{M}\Tr\bigl[\G\n \Jtrue\bigr] = \frac{1}{M}\Tr\Bigl[\frac{\Jtrue}{\I+\Lambda \Cpop}\Bigr] 
  = \frac{1}{M}\Tr\Bigl[\frac{\Jtrue^2}{\Jtrue+\Lambda }\Bigr]
\]
For the first term we use the following identity:
\[
{\G\n}^2 = \G\n)+\alpha\frac{d}{d\alpha} \G\n 
\]
As a result, asymptotically we have:
\begin{align*}
  \Tr\Bigl[{\G\n}^2\Bigr] &= \Tr\Bigl[\frac{1}{\I+\Lambda \Cpop}\Bigr]+\alpha\frac{d}{d\alpha} \Tr\Bigl[\frac{1}{\I+\Lambda \Cpop}\Bigr] \\[0.2cm]
  &= \bigl(1-\Lambda'(\alpha)\bigr)\Tr\Bigl[\frac{1}{\I+\Lambda \Cpop}\Bigr]+\Lambda'(\alpha) \Tr\Bigl[\frac{1}{(\I+\Lambda \Cpop)^2}\Bigr]
\end{align*}
For this we need to compute $\Lambda'(\alpha)$ which can be done from the self-consistent equation~(\ref{eq:rmt_lambda},\ref{eq:rmt_gamma}):
\[
\Lambda'(\alpha) = \frac{\Lambda^2}{\alpha^2}\frac{\rho}{\rho-Q[\Lambda]}
\]
with
\[
Q[\Lambda] = \frac{1}{M}\Tr\Bigl[\frac{\Lambda^2 \Cpop^2}{(\I+\Lambda \Cpop)^2}\Bigr]
\]
Ultimately we obtain:
\[
\mathcal{E}_J = \frac{1}{M}\Tr\Bigl[\Bigl(\frac{\alpha}{1+\Lambda \Cpop}-\frac{1}{\Cpop}\Bigr)^2\Bigr]+\frac{\alpha^2(1-\Lambda')}{M}\Tr\Bigl[\frac{\Lambda \Cpop}{(1+\Lambda \Cpop)^2}\Bigr],
\]
So we have
\[
\mathcal{E}_J = \int\nu(dx)\bigl[\frac{\alpha}{1+\Lambda x}-\frac{1}{x}\bigr]^2+\alpha^2(1-\Lambda')\int\nu(dx)\frac{\Lambda x}{(1+\Lambda x)^2}
\]
Finally, concerning $LL_{\rm train,test}$ we don't see how to avoid the Cauchy integral, 
but the train-test relationship~(\ref{eq:train-test}) has an important consequence, 
because it allows us to get a very precise estimation of the test likelihood when $M$ becomes large:
\[
LL_{\rm test}(J)  =  \frac{1}{2}\log\det(J) - \frac{E_{\rm train}}{1-\frac{1}{\rho}E_{\rm train}} 
\]
as long as  $J$ is the function~(\ref{eq:j_L1}) of $C\n$. For general EBM models we have a LL of the form
\[
LL_{\rm train,test}[J] = -\log Z[J] - E_{\rm train,test}[J]
\]
so by analogy with GCV, it is not excluded that we can use this train-test relation in practice.  

\newpage
\section{Details on data-correction protocols \label{app:data_correction_details}}

In this section, we analyze the effect of different ways to improve the estimation of the covariance matrix's eigenvalues in order to avoid or diminish the effect of overfitting during the training dynamics.

\subsection{Training dynamics with regularization prior for finite $N$ \label{subapp:regularization}}

We first discuss what happens to the training in the presence of a regularization. We employ two regularization protocols: a standard $L_2$-norm, and a projected $L_1$-norm. The choice of the second regularization is justified because it allows to have a maximum-a-posteriori coupling matrix which commutes with the original covariance matrix $\CM$, as it happens in the absence of regularization, thus facilitating the asymptotic analysis through RMT discussed in Appendix~\ref{sec:RMT}. 
\subsubsection{$L_2$ regularization} The log-posterior now reads 
\begin{equation}
    \frac{1}{M}\log p\left(\bm J \mid \mathcal{D}\right) = \like_{\mathcal{D}} \paren{\bm J} - \frac{\lambda}{4} \text{Tr} \left( \bm J ^2 \right)
\end{equation}
where $\lambda$ is the reguavlarization strength. The derivative w.r.t. the parameters now reads 
\begin{equation}
    \frac{1}{M}\frac{\partial \log p\left(\bm J \mid \mathcal{D}\right)}{\partial J_{ij}} = \left[  - \hat{C}^M_{ij} + \left(\bm J^{-1}\right)_{ij} -\lambda J_{ij} \right]\label{eq:gradient_gaussian_model_withL2}
\end{equation}
Notice that the new term commutes with the second one ($\bm J$ and $\bm J^{-1}$ are diagonal in the same basis), so even in this case the maximum-a-posteriori matrix $\widehat{\bm J}^{\mathrm{MAP}}$ will share the same basis as $\CM$, as it happens in the absence of regularization. Therefore, we can apply the same reasoning discussed in the main text and project the log-posterior's gradient on the basis of $\bm J$. The evolution equation of each eigenvalues reads:
\begin{equation}
    \tau\frac{\der J_\alpha}{\der t} = \frac{1}{J_\alpha}-\CMval - \lambda J_{\alpha}, \label{eq:grad_Ja_withL2}
\end{equation}
Although there exist no closed expression for the full time-dependent solution of Eq.~\eqref{eq:grad_Ja_withL2}, it is possible at least to compute analytically its fixed point:
\begin{equation}
J_{\alpha}^{(\infty)-L_2} \paren{\lambda} = \frac{1}{2\lambda}\left[- \CMval + \sqrt{\paren{\CMval}^2 + 4 \lambda }\,\right] \label{eq:Ja_fp_L2}
\end{equation}
The full coupling matrix corresponding to the above fixed point is finally computed projecting back Eqs.~\eqref{eq:Ja_fp_L2}
 onto the eigenbasis of $\CM$, that is $
\bm{J}^{(\infty)-L_2}  \paren{\lambda} = \sum_{\alpha} J_{\alpha}^{(\infty)-L_2} \paren{\lambda} \CMvec \CMvec ^\top 
$.

\begin{figure*}[h!]
\begin{overpic}[width=\textwidth]{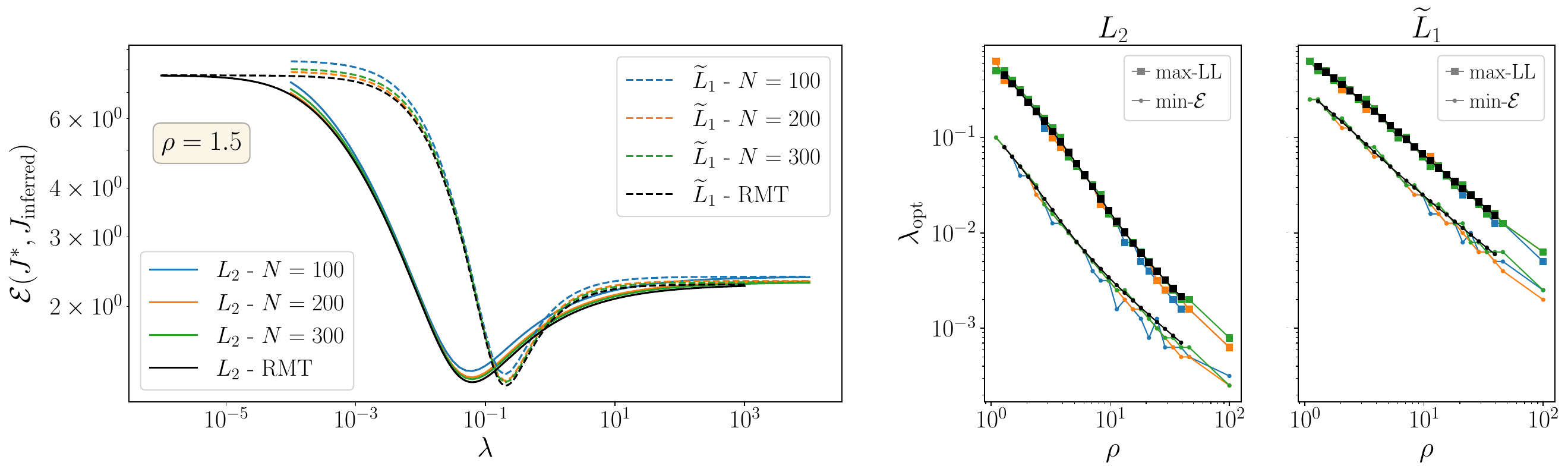}
\put(3,28){{\figpanel{a}}}
\put(60,28){{\figpanel{b}}}
\put(80,28){{\figpanel{c}}}
\end{overpic}
\caption{Effect of the regularization priors on the inferred model's quality. \figpanel{a}: the plot shows the reconstruction error  $\mathcal{E}_{\text{J}}$ computed between the ground truth and the inferred model in the presence of a regularization prior with strength $\lambda$. Solid lines refer to the $L_2$ norm, while dashed lines to the spectral-$L_1$ norm, both discussed in Section~\ref{subapp:regularization}. For each prior, we compare finite-size results (colored lines) and RMT asymptotic estimations. 
\figpanel{b}-\figpanel{c}: we show the values of the regularization strength $\lambda_{\text{opt}}$ that achieves optimal reconstruction of the model (lines with scatter points), and the optimal value of the regularization that maximizes the test log-likelihood (lines with scatter diamonds). Panels \figpanel{b} (resp. \figpanel{c}) refers to the optimal values when using the $L_2$-norm (resp. the spectral $L_1$-norm). Note that \figpanel{b}-\figpanel{c} are on the same y-scale. Settings are the same as in Fig.~\ref{fig:rmt_comparison_dynamics}. \label{fig:comparisonL2L1appendix}}
\end{figure*}

\subsubsection{Spectral $\widetilde{L}_1$-norm \label{sec:RMT_regularizationL1}}
This regularization schemes utilizes a $L_1$-norm but on the projected basis of the coupling matrix $\bm J$. This construction still allows to employ a similar formula to Eq.~\eqref{eq:Jalpha_lambert_Ising}
 to describe the evolution of eigenvalues, each one independently on the others. The original differential equation describing the evolution of $J_{\alpha}$ is now modified as 
 \begin{equation}
    \frac{1}{\gamma}\frac{\der J_\alpha}{\der t} = \frac{1}{J_\alpha}-\CMval - \lambda , \label{eq:grad_Ja_withL1_modes}
\end{equation}
whose fixed point reads \begin{equation}
J_{\alpha}^{(\infty)-\widetilde{L}_1} \paren{\lambda} = \frac{1}{\CMval + \lambda}\label{eq:Ja_fp_L1_modes} 
\end{equation}
Note that equations ~\eqref{eq:Ja_fp_L2}-~\eqref{eq:Ja_fp_L1_modes} just derived 
 are the same ones as Eqs.~\eqref{eq:j_L2}-\eqref{eq:j_L1}  in Appendix ~\ref{sec:BM}, respectively. 
 \subsubsection{Results on the effect of regularization}
We can check the performances of either type of regularization by looking at the training fixed point, and at how it modifies the quality of the inferred model. In Fig.~\ref{fig:comparisonL2L1appendix}-\figpanel{a},  we show the reconstruction error $\mathcal{E}_{\text{J}}$ computed between the ground truth and the inferred model in the presence of a regularization prior with strength $\lambda$, both for the $L_2$-norm (solid lines) and for the spectral-$L_1$ norm (dashed lines), for a given value of number of samples: here we have $\rho = M/ N = 1.5$. Comparison is shown between finite-size trainings (colored lines) and RMT estimation (black lines). All the quantities are plotted versus the regularization strength $\lambda$: we can observe how there is clear non-monotonic behavior with a minimum developing at a certain $\lambda_{\text{opt}}$. As one might expect the optima value differs between the two regularization priors, i.e. $\lambda^{L_2}_{\text{opt}} \neq \lambda^{\widetilde{L}_1}_{\text{opt}}$. Results on both priors are shown in the same panel to highlight that the two regularization schemes have qualitatively the same effect on the final reconstruction error: that is, the quality of the inferred model at the optimal value is the same for both regularizations.
Panels \figpanel{b}-\figpanel{c} show instead the optimal value of $\lambda$ computed either by minimizing the reconstruction error (shown with points) or by maximizing the test log-likelihood (diamonds). Panel \figpanel{b} is actually a repetition of Fig.~\ref{fig:rmt_comparison_dynamics}-\figpanel{d} and refers to the $L_2$-prior, while \figpanel{c} refers to the $L_1$ spectral prior. Again, we can observe a similar behavior of the two norms, with the only difference that $\lambda^{L_2}_{\text{opt}} < \lambda^{\widetilde{L}_1}_{\text{opt}}$ independently on the chosen criterion. Finally, all the optimal values go to $0$ when $\rho \to \infty$, as expected.

\begin{figure*}[t!]
 \centering
\begin{overpic}[width=.8\textwidth]{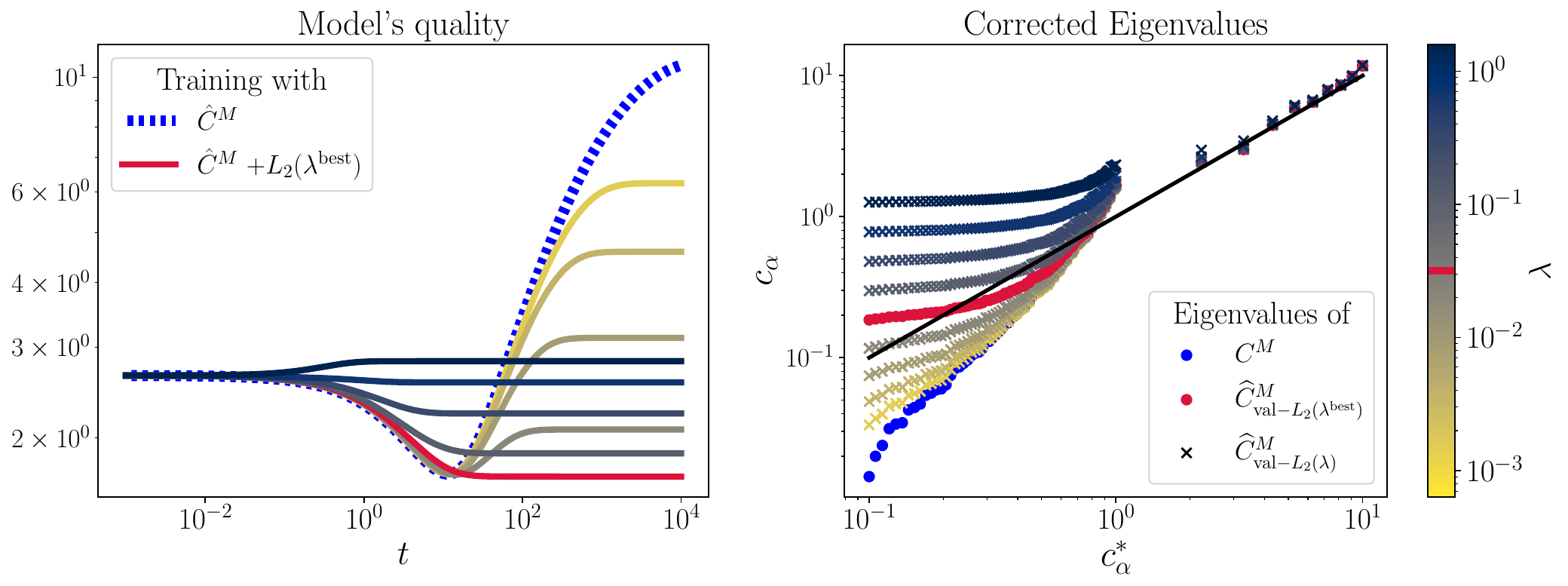}
\put(7,21.7){{\figpanel{a}}}
\put(55.9,30){{\figpanel{b}}}
\end{overpic}
\caption{Effect of different $L_2$-norm regularization strengths $\lambda$ on the GEBM's learning dynamics. Panel \figpanel{a} shows the reconstruction error vs time. The dotted blue line corresponds to the standard training over $\CM$. All the other full lines correspond to a training with a certain value of regularization strength $\lambda$, obtained by numerically solving Eq.~\eqref{eq:grad_Ja_withL2} for all modes. The regularization strength $\lambda$ increases from yellowish colors to blueish (see the colorbar at the right). The curve corresponding to the optimal regularization that minimizes the reconstruction error after training is highlighted in red. Panel \figpanel{b}: plot of the equivalent covariances modes corrected by the regularization. For each curve, we scatter plot these values Eq.~\eqref{eq:Cval_L2corrected_def} against the population eigenvalues. Here we set $\rho=1.66$. \label{fig:gaussian_model_regularization_effect_on_dynamics}}
\end{figure*}

What is the effect of the regularization on the training dynamics? Considering that the standard training has a non-monotonic behavior w.r.t. the training time, we would expect that, since the regularization strongly improves on the models' quality w.r.t. the standard case (at least at the optimal optimizing regularization strength $\lambda_{\mathrm{opt}}$), such a non-monotonic behavior is diminished. This is indeed the case, as shown by Fig.~\ref{fig:gaussian_model_regularization_effect_on_dynamics}-\figpanel{a}, displaying different training curves for different regularization strengths: the closest the regularization to its optimal value (highlighted in red), the smoother the model's quality is w.r.t. training time. At the optimal point the model's quality is completely non-monotonic and approaches the fixed point at the same reconstruction error as the minimum w.r.t. time.  

Actually, due to the simplicity of the GEBM it is even possible to interpret the $L_2$ regularization as a shrinkage correction protocol. Consider indeed the training fixed point given by Eq.~\eqref{eq:grad_Ja_withL2}. We stress again that the maximum-posterior matrix $\bm J$ has the same basis decomposition as the empirical covariance matrix, because the regularization term commutes with the other two terms in Eq.~\eqref{eq:grad_Ja_withL2}.
By the dualism between covariance matrix and coupling matrix in the Gaussian EBM, we can think at the reciprocal values of Eq.~\eqref{eq:grad_Ja_withL2} as eigenvalues of a corrected covariance matrix w.r.t. $\CM$, depending on $\lambda$. We can therefore define another eigenvalue-corrected covariance matrix, by using the analytic fixed point on the training dynamics obtained through the regularization:
\begin{equation}
\CCorrectedRegu = \sum_{\alpha} \frac{1}{J_{\alpha}^{(\infty)-L_2} \paren{\lambda} } \CMvec {\CMvec}\,\!^{\top} \label{eq:Cval_L2corrected_def}\\
\end{equation}
By definition, the training fixed point obtained using $i)$ a training dynamics with the un-touched empirical covariance matrix $\CM$ plus the regularization term or $ii)$ a regularization-free dynamics using matrix \eqref{eq:Cval_L2corrected_def} are the same. Fig.~\ref{fig:gaussian_model_regularization_effect_on_dynamics}-\figpanel{a} shows indeed how the regularization modifies the eigenvalues of $\CM$ when interpreting the fixed point of the training dynamics as a shrinkage correction. Each set of points shows the quantities $1 \slash J_{\alpha}^{(\infty)-L_2} \paren{\lambda} $ (i.e. the eigenvalues of \eqref{eq:Cval_L2corrected_def} ) vs the population ones $\Cpopval$. It is intuitive to notice that the optimal regularization (red points) is the one that makes such corrected eigenvalues as close as possible to the population ones. This entire reasoning holds analogously with the spectral $L_1$ norm. 

\subsection{Empirical shrinkage correction through modes fitting \label{subapp:fit}}
\begin{figure}
\begin{overpic}[width=0.45\textwidth]{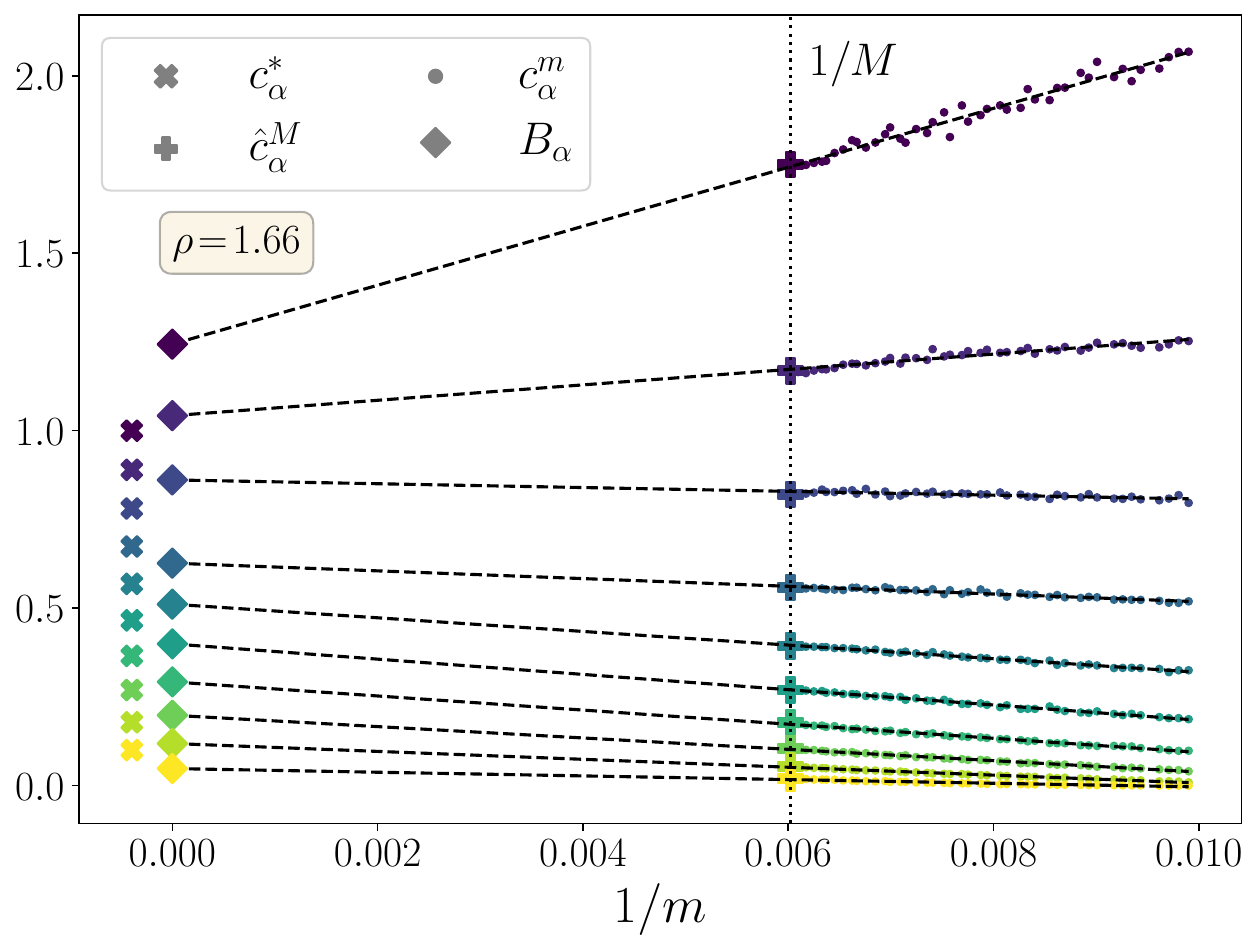}
\put(-2,73){{\figpanel{a}}}
\end{overpic}    
\hfill
\begin{overpic}[width=0.45\textwidth]{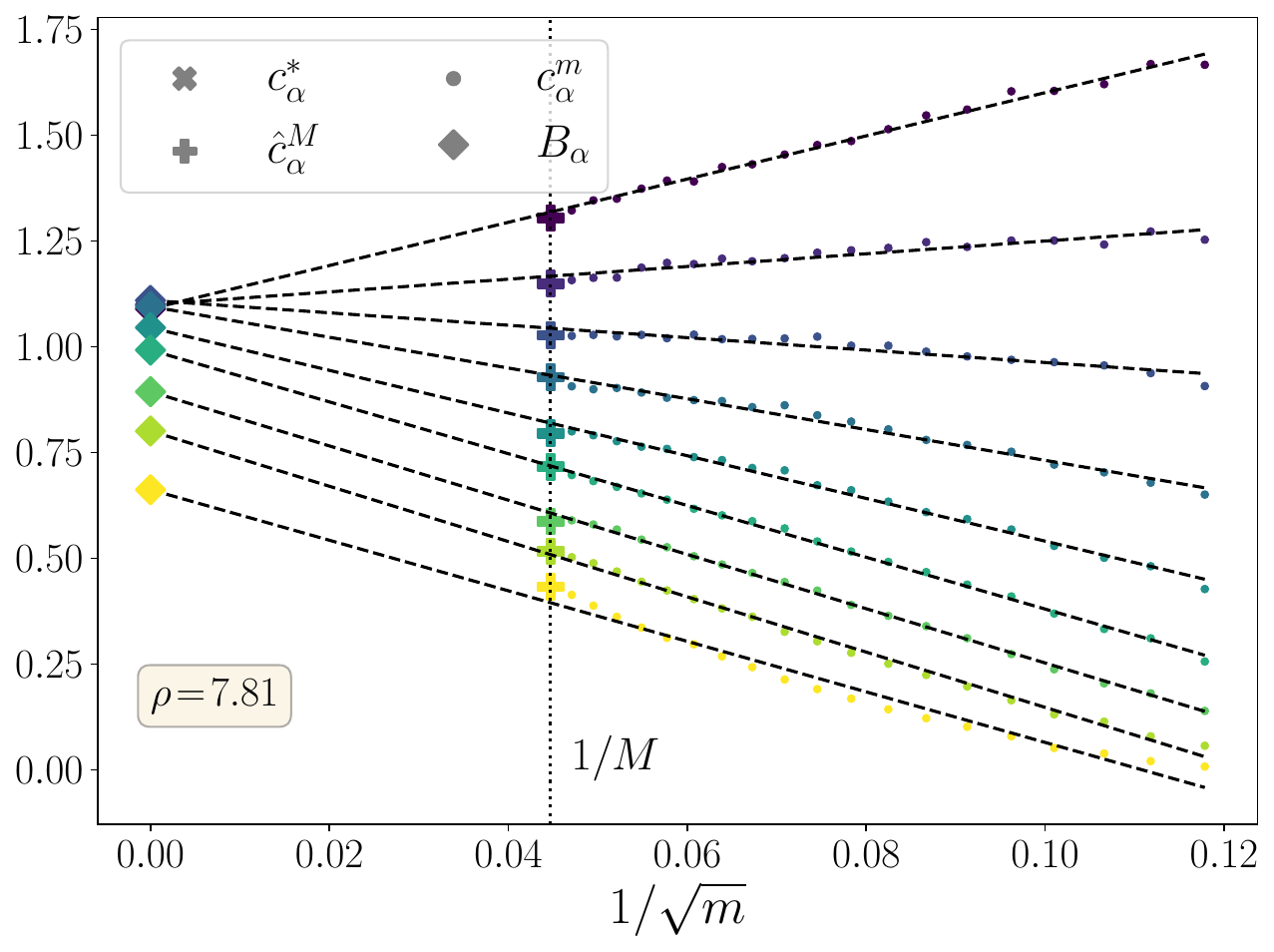}
\put(-5,73){{\figpanel{b}}}
\end{overpic}    
\caption{\textbf{Examples of eigenmode fitting procedures for the GEBM and the Ising-BM.} Each panel illustrates the procedure used to fit the eigenmodes of the covariance matrix $\CM$ by downsampling to $m < M$ samples, in order to extrapolate their behavior as $m \to \infty$, following Eq.~\eqref{eq:fit_equation}.
Panel~\figpanel{a} refers to the GEBM used in the main text (e.g., Fig.~\ref{fig:datacorrection}). For a fixed value of $M$ such that $\rho = M/N = 1.66$, we show a subset of eigenvalues $\{\CMval\}$ (denoted by + markers), along with their downsampled counterparts $\{c_\alpha^m\}$ for several values of $m < M$ (small circles). These downsampled eigenvalues are obtained by randomly selecting $m$ samples from the full dataset and computing the eigenvalues of the resulting covariance matrix, averaged over 10 independent instances.
Dashed black lines correspond to fits using Eq.~\eqref{eq:fit_equation} with $\nu = 1$, and colored diamonds indicate the extrapolated intercepts $B^M_\alpha$, i.e., the estimated eigenvalues at $m \to \infty$. Crosses mark the population eigenvalues $\{\Cpopval\}$, showing that the fitted extrapolations are significantly closer to the true population spectrum than the empirical eigenvalues obtained from $M$ samples.
Panel~\figpanel{b} shows the analogous procedure for the Ising-BM, using $\rho = 7.81$. In this case, the best fits are obtained with $\nu = 1/2$, and the horizontal axis is accordingly rescaled as $1/\sqrt{m}$.}
    \label{fig:example_fit}
\end{figure}

A simple way to perform a heuristic shrinkage correction is to down-sample the empirical covariance matrix and estimate the asymptotic eigenvalues through a fitting procedure. Starting from the available dataset with $M$ samples - whose covariance matrix is $\CM$ - we can randomly extract 
subsets of $N<m<M$ samples and estimate the eigenvalues of the size-reduced covariance matrices $\widehat{\bm C}^m$. Every time a down-sampling procedure of this kind is performed, both the eigenvalues and the eigenvectors will be different from the original $\CM$; however, we here suppose to account for the eigenvalues, keeping the basis fixed to the one of $\CM$. After applying this computation to different values of $N<m<M$, for each eigenvalue $\alpha$ we can fit the resulting data $\lazo{\hat{c}^m_{\alpha}}_{m\in (N; M]}$ according to \begin{equation}
    \hat{c}_{\alpha} \left(m\right) = \frac{1}{m^{\nu}} A^M_{\alpha} + B^M_{\alpha} \label{eq:fit_equation}
\end{equation}
with $\nu$ being an exponent of choice. The coefficients $B^M_{\alpha}$ will represent the asymptotic estimate of the $\alpha$-th eigenvalue of the population matrix, $\Cpopval$, corresponding to the $m\to\infty$ extrapolation. After fitting each mode separately, we can construct an eigenvalue-corrected covariance matrix as
\begin{equation}
\widehat{\bm C}_{\mathrm{val}-\mathrm{fit}} = \sum_{\alpha} B_{\alpha}^M \CMvec {\CMvec}\,\!^{\top} \label{eq:Cval_fit_corrected}\\
\end{equation}
This procedure can in principle be generalized e.g by using a combination of powers in the fitting function, although in this work we only restricted to the functional form  \eqref{eq:Cval_fit_corrected}; secondly, the estimation of the fitting coefficients can be improved by collecting mean values of eigenvalues  $\lazo{\hat{c}^m_{\alpha}}_{m\in (N; M]}$ by evaluating the down-sampled covariance matrix $\widehat{\bm C}^m$ multiple times with different subsets of the $M$ original data.
In the experiments presented in the main text for the GEBM (orange line in Fig.~\ref{fig:datacorrection}-\figpanel{b}), we used a simple linear fitting in $1\slash m$ ($\nu=1$) and $10$ random resampling for each value of $m$, from which the mean eigenvalue is extracted to perform the fit. This linear scaling (in $1\slash m$) for the finite-size fluctuation of the eigenvalues in the GEBM is also justified by theoretical evidence (see e.g. \cite{ledoit2011eigenvectors}).
In the experiments for the Ising-BM instead, (orange line in Fig.~\ref{fig:BM}-\figpanel{c}) we find that best reconstruction is achieved with $\nu=1 \slash 2$, while a linear fitting has always very bad performances. Also in this case we used $10$ resampling steps. 
An example of such a fitting procedure is shown in Figure~\ref{fig:example_fit} for both the GEBM (in \figpanel{a}) and for the BM (in \figpanel{b}), in each case for a given value of $M$.

\section{Eigendecomposition of training dynamics on Boltzmann Machine \label{app:BMcalculations}}
We consider a Ising-like Boltzmann Machine for the inference of binary-valued data. The probability of a configuration $\bm x$ where $x_i \in [-1,1]$ at given parameters is expressed as
\begin{equation} 
p \paren{\bm x \mid \bJ, \bm h}=\frac{1}{Z} e^{\sum_{i<j} J_{ij} x_i x_j + \sum_i x_i h_i}.\label{eq:Isingmodel}
\end{equation}
We suppose to generate equilibrium configurations from a known model with $ {\bm \theta}^*  = \paren{{\bm J}^*, {\bm h}^*}$ (eventually these parameters are rescaled by an external factor $\beta$ that plays the role of an external inverse temperature), and we want to infer back the original model through a likelihood maximization procedure. The LL of a certain set of parameters $\bm \theta  = \paren{\bm J , \bm h}$ is given by
\begin{equation}\label{eq:gradient}
    \mathcal{L}_{\mathcal{D}} \paren{\bm J, \bm h} = \frac{1}{M} \sum_{\mu=1}^{M} \log p \paren{\bm x^\mu \mid \bJ, \bm h} = \sum_{i<j} J_{ij} \mean{x_i x_j}{\mathcal{D}} + \sum_i h_i \mean{x_i}{\mathcal{D}} - \log Z
\end{equation}
where $\mean{\cdot}{\mathcal{D}}$ denotes the average w.r.t. the dataset $\mathcal{D} = \left\{ \bm x_\mu \right\}_{\mu=1}^M$. In what follows, for the analytic treatment of the training dynamics in the ML procedure we neglect the problem of learning the external fields $h_i$. This assumption is consistent with a scenario where the data have null magnetizations. Therefore, the gradient of the LL w.r.t. the couplings $J_{ij}$ reads
\begin{equation}\label{eq:gradient}
    \frac{\partial \like_{\mathcal{D}}}{\partial J_{ij}}=\mean{x_i x_j}{\pemp}-\mean{x_i x_j}{\bm J} =\CM-\mean{x_i x_j}{\bm J},
\end{equation}
Then, the couplings are updated as
\begin{equation}
    J_{ij}(t+1)\leftarrow J_{ij}(t)+\gamma\frac{\partial \like_{\mathcal{D}}(t)}{\partial J_{ij}},
\end{equation}
where $\gamma$ is the learning rate. From here, we can assume an ideal training with an infinitesimal learning rate to recast the evolution equation of the matrix $\bm J$ in time in the following matrix form
\begin{equation}
    \tau\frac{d J_{ij}}{dt} =\left.\frac{\partial \like_{\mathcal{D}}}{\partial J_{ij}}\right|_{J(t)} \qquad \qquad \Longrightarrow \qquad \qquad
    \tau\frac{d \bm J}{dt}=\CM- \av{\bm x \bm x^\top}_{\bm J}. \label{eq:grad_original}
\end{equation}
where $\mean{\cdot}{\bm J} $ denotes the average w.r.t. to model \eqref{eq:Isingmodel}, and $\tau=1/\gamma$. Note that, since we assumed to neglect local magnetizations, the r.h.s \eqref{eq:grad_original} computed in its diagonal entries is $0$. This is consistent with the fact that self-couplings $J_{ii}$ do not evolve in time, because they correspond to constant energy terms in the energy function. In order to make the models' correlation $\mean{\bm x \bm x^\top}{\bm J}$ analytically treatable, we implement now a mean-field approximation. 
We can exploit the following exact expression (self-consistent) for the correlator \cite{suzuki1968dynamics}, which we further expand for high temperatures
\begin{equation}
    \mean{x_i x_j}{\bm J}=\delta_{ij}+(1-\delta_{ij})\mean{x_i \tanh{ \sum_{k}J_{jk}x_k}}{\bm J }\approx \delta_{ij}+(1-\delta_{ij})   \sum_{k}J_{jk} \mean{x_i x_k}{\bm J},
\end{equation}
Then, in matrix form ($C_{ij}=\mean{x_i x_j}{\bm J})$ we get
\begin{equation}
    \bm C=\mathbb{I}_N + \bm J \bm C - \text{diag}\caja{\bm J \bm C}\label{eq:CIsing_MF}
\end{equation}
where $\mathbb{I}_N$ is the identity matrix of size $N$, and the operator $\text{diag}\caja{\mathcal{M}}$ extracts the diagonal part of a matrix $\mathcal{M}$. The last term is introduced to correct the wrong estimation of the diagonal entries of $\bm C$, which should be equal to $1$. The problem is that the above equation does not admit a simple analytical solution for the model's correlation matrix $\bm C$, which was the original goal. Instead,  is typical implemented in the literature is the following expression of the linear response correlations
\begin{equation}
    \bm C=f(\bm J)=( \mathbb{I}_N -\bm J)^{-1} \label{eq:CMFapproximate}
\end{equation}
which gives a reliable estimate of the correlation matrix and is at the core of well-studied mean-field like expression for the inferred couplings \cite{kappenrodriguez_MFBM, Ricci-Tersenghi_2012}. However, the diagonal entries of \eqref{eq:CMFapproximate} are not in general equal to $1$. Normally, this is not an issue because one is interested in correlations for $i\neq j$ (i.e. off-diagonal entries). However, in our approach such a diagonal mismatch creates a non-null gradient on the diagonal entries of $\bm J$. Indeed, by plugging Eq.~\eqref{eq:CMFapproximate} into the LL's gradient, we get
\begin{equation}
    \tau\frac{d \bm J}{dt}=\CM-(\mathbb{I}_N -\bm J)^{-1} \label{eq:gra_MF_ising}
\end{equation}
Now, the diagonal part of the r.h.s. of Eq.~\eqref{eq:gra_MF_ising} not null anymore. In order to circumvent this additional issue, a possible solution could be to modify the gradient by remove the diagonal terms -  which would results in an nonphysical evolution of the self-couplings - "by hand" : 
\begin{equation}
    \tau\frac{d \bm J}{dt}=\CM-(\mathbb{I}_N -\bm J)^{-1}  + \text{diag}\left[   \mathbb{I}_N  - (\mathbb{I}_N-\bm J)^{-1} \right] \label{eq:CIsing_MF_spherical}
\end{equation}
The last term in the above expression correctly fixes the diagonal problem for the matrix $\bm J$ in the gradient. This strategy is similar to what carried out in \cite{Fanthomme_2022} where the authors impose a add a spherical constraint on the gradient in the form of a Lagrange multiplier. However, this leads to a complicated expression even for the training fixed point because the evolution of all the eigenvalues is now coupled. For this reason, since here we are interested in the dynamics of training, adding this constraint would result  into a system of coupled differential equations for the eigenvalues, which has computationally the same complexity of the original problem, so there would be no gain in that. 
The simplest strategy is therefore to use the approximate expression for the correlator and avoid adding the constraint, so to use the gradient \eqref{eq:gra_MF_ising}  as it is. Although it might seem a crude approximation, it still allows us to decompose the dynamics in the same way as for the GEBM. Before going on, it is worth noticing that the diagonal matching problem is at the core of some refined mean-field like approximations for binary (Ising-like) maximum-entropy models which exploit e.g. iterative diagonal consistency tricks (see e.g \cite{diagonal_consistency_tanaka,diagonal_matching_kiwata} ).  
Therefore, as explained in Sec.~\ref{sec:training} for the GEBM we can project the log-likelihood's gradient onto the eigenbasis of $\bm J$, leading to an expression for the rotation of eigenvectors (same one as in \eqref{eq:grad_Ja}) and another one for the evolution of its eigenvalues, which is given below: 
\begin{equation}
    \!\!\tau\frac{\der J_\alpha}{\der t} = \hat{c}_{\alpha \alpha}^M - \frac{1}{ \mathbb{I}_N-J_\alpha}
\end{equation}
which is the same equation shown in the main text (Eq.~\eqref{eq:grad_Ja_ising}). 
As explained in the main text, the solution of the above equation describes an independent evolution of eigenvalues which is not quantitative accurate with respect to the numerical results, but still captures the qualitative trend: in particular, it perfectly describes the separation of timescales in terms of PCA's modes during the training dynamics, an effect which is observed also in the numerical results (see panels \figpanel{b} in Fig.~\ref{fig:BM}). 

\begin{figure*}[h!]
\centering
\begin{overpic}[width=.8\textwidth]{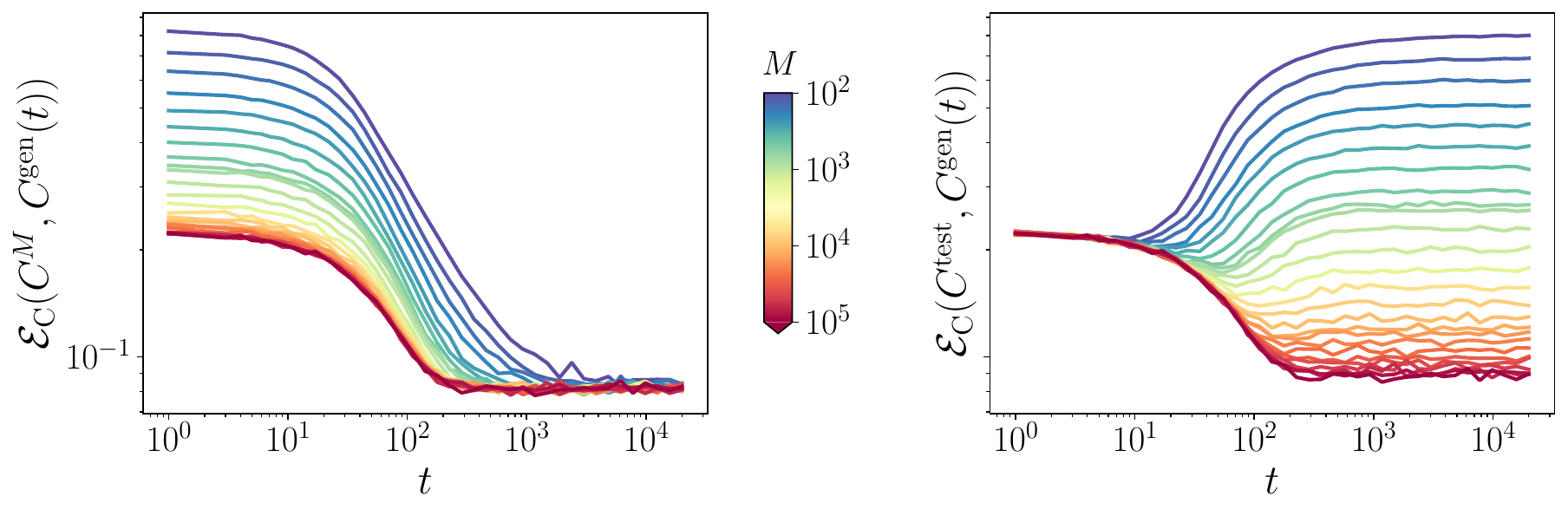}
\put(5,31){{\figpanel{a}}}
\put(58,31){{\figpanel{b}}}
\end{overpic}
\caption{Supplementary results on the Boltzmann Machine for the inverse Ising problem. The model, dataset and training setting are the same as in Fig.~\ref{fig:BM}. \figpanel{a}: we show the error between the covariance matrix of generated configurations from the model (along the training trajectory) and the covariance matrix of the training set $\CM$, so $\mathcal{E}_{\CM} = \left \lVert \CM - \bm C^{\text{gen}} (t) \right \rVert_\text{F} $. \figpanel{b}: plot of the generation error computed w.r.t. a test set, again between the covariance matrices, i.e. $\mathcal{E}_{\bm C^{\text{test}}} = \left \lVert \bm C ^{\text{test}} - \bm C^{\text{gen}} (t) \right \rVert_\text{F} $. Both quantities are plotted versus training time (number of updates) for different values of $M$ shown in the colorbar. The learning rate is set to $\gamma=10^{-2}$.\label{fig:BM_generationerror}}
\end{figure*}

\begin{figure*}[t!]
\centering
\begin{overpic}[width=.6\textwidth]{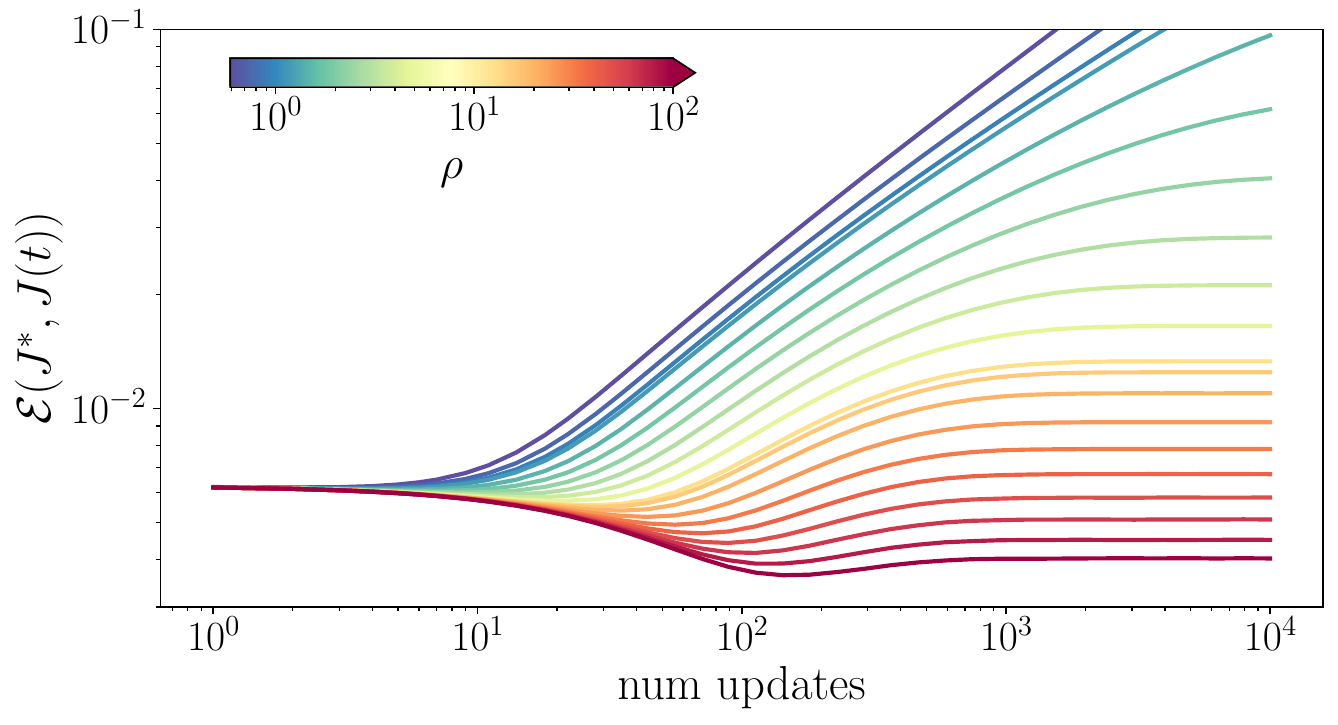}
\end{overpic}
\caption{Supplementary results on the Boltzmann Machine for the inverse Ising problem. The model, dataset and training setting are the same as in Fig.~\ref{fig:BM}, except for the system size which here is equal to a lattice size of $L=32$, so that $N=L^2=1024$.  We show the Frobenius norm of the error between the true model and the trained one versus training time (number of updates) for different values of $\rho=M \slash N$ shown in the colorbar. The learning rate is set to $\gamma=10^{-2}$.\label{fig:BM_larger_L32}}
\end{figure*}
\end{document}